%% file: main.tex
\documentclass{article}
\usepackage[preprint,nonanonymous]{neurips_2026}
\usepackage[utf8]{inputenc} 
\usepackage[T1]{fontenc}    
\usepackage{hyperref}       
\usepackage{url}            
\usepackage{booktabs}       
\usepackage{amsfonts}       
\usepackage{nicefrac}       
\usepackage{microtype}      
\usepackage[table]{xcolor}  
\usepackage{placeins}
\usepackage[utf8]{inputenc}
\usepackage[T1]{fontenc}
\usepackage{hyperref}
\setcitestyle{numbers,square}
\definecolor{tocblue}{RGB}{31, 73, 125} 
\hypersetup{
    colorlinks=false,
    citebordercolor=green,
    linkbordercolor=green,
    urlbordercolor=blue,
    pdfauthor={Yangshuang Xu, Yuyang Dai, Liling Chang, Qi Wang, Yushun Dong},
    pdftitle={Does Your Wildfire Prediction Model Actually Work, or Just Score Well?},
    pdfsubject={},
    pdfkeywords={}
}

\usepackage{url}
\usepackage{booktabs}
\usepackage{amsfonts}
\usepackage{nicefrac}
\usepackage{microtype}
\usepackage{amsmath}
\usepackage{amssymb}
\usepackage{graphicx}
\usepackage{tabularx}
\usepackage{longtable}
\usepackage{multirow}
\usepackage{array}
\usepackage{float}
\usepackage{adjustbox}
\usepackage{placeins}
\usepackage{enumitem}
\usepackage{siunitx}
\usepackage{tikz}
\usepackage{subcaption}
\usepackage{wrapfig}
\usepackage[normalem]{ulem}
\usepackage{pifont}
\usepackage{hyperref}
\usepackage{xcolor}
\usepackage{tabularx}
\usepackage{xspace}
\usepackage{fontawesome5}

\definecolor{wfblue}{RGB}{42,111,151}
\definecolor{wforange}{RGB}{231,111,81}
\definecolor{wfgreen}{RGB}{42,157,143}
\definecolor{wfgold}{RGB}{233,196,106}
\definecolor{wfslate}{RGB}{38,70,83}
\definecolor{wfgray}{RGB}{108,117,125}
\definecolor{wfpurple}{RGB}{116,81,164}
\definecolor{wfindigo}{RGB}{77,100,166}
\definecolor{wfrose}{RGB}{188,80,144}
\definecolor{wfolive}{RGB}{120,143,64}
\definecolor{primarybg}{RGB}{219,234,254}
\definecolor{primaryrule}{RGB}{147,197,253}
\definecolor{headerbg}{RGB}{30,58,138}
\definecolor{regbg}{RGB}{240,240,240}
\definecolor{retrbg}{RGB}{232,232,232}
\definecolor{refbg}{RGB}{255,243,205}      
\definecolor{alphabg}{RGB}{220,237,220}    
\definecolor{subheadbg}{RGB}{241,245,249}  
\definecolor{bestval}{RGB}{0,100,0}        
\definecolor{warnval}{RGB}{180,0,0}        

\newcolumntype{L}[1]{>{\raggedright\arraybackslash}p{#1}}
\newcolumntype{Y}{>{\raggedright\arraybackslash}X}
\newcommand{\ms}[2]
{\ensuremath{#1{\mkern1mu}_{\scriptscriptstyle \pm #2}}}

\newcommand{\ourfm}{\textsc{Wild}{\textbf{FIRE}}\textsc{-FM}\xspace}
\title{Does Your Wildfire Prediction Model Actually Work,\\ or Just Score Well?}

\author{%
  Yangshuang Xu\thanks{Equal contribution.} \\
  Florida State University \\
  \texttt{yx21e@fsu.edu} \\
  \And
  Yuyang Dai\footnotemark[1] \\
  Florida State University \\
  \texttt{yd26@fsu.edu} \\
  \And
  Liling Chang \\
  Florida State University \\
  \texttt{liling.chang@fsu.edu} \\
  \And
  Qi Wang \\
  Northeastern University \\
  \texttt{wangqi@vt.edu} \\
  \And
  Yushun Dong\thanks{Corresponding author.} \\
  Florida State University \\
  \texttt{yd24f@fsu.edu} \\
}

\begin{document}

\maketitle

\input{sections/0_abstract}

\begin{center}
\small
{\large\faDatabase}~\href{https://huggingface.co/RAI-Lab/Wildfire-FM}{\texttt{https://huggingface.co/RAI-Lab/Wildfire-FM}}
\end{center}

\input{sections/1_intro}
\input{sections/2_backbone}
\input{sections/3_prelim}
\input{sections/4_experiments}
\input{sections/5_conclusion}

\bibliographystyle{plain}
\bibliography{references}

\newpage
\input{sections/appendix}

\end{document}

%% file: sections/0_abstract.tex
\begin{abstract}
Wildfire prediction is important for early warning and resource allocation, yet existing Earth foundation models (Earth FMs) are pretrained for general atmospheric and geophysical objectives rather than wildfire forecasting. To address this gap, we introduce \ourfm, the first foundation model pretrained specifically for wildfire prediction using weather, active-fire observations, topography, vegetation, and static environmental data. However, introducing a domain-specific backbone alone does not solve the evaluation problem: wildfire events are sparse in space and time, making transfer conclusions highly sensitive to matching rules and evaluation settings.
To address this problem, we introduce a fixed-contract evaluation framework with two controlled checks: a fixed-output check for matching-rule effects and a fixed-feature check for head-selection effects. Under matched contracts, we compare \ourfm\ with ten Earth-FM baselines across occupancy, spread, retrieval, and regression tasks. Our results show that wildfire transfer conclusions depend strongly on evaluation design and task formulation. We hope this framework and \ourfm\ provide a foundation for future wildfire-specific Earth-FM research and benchmarking.
\end{abstract}

%% file: sections/1_intro.tex
\section{Introduction}

Wildfire prediction is critical for early warning and resource allocation in disaster response~\cite{goldammer1999early, farahmand2020fdeo}. As extreme fire events grow more frequent and severe, accurate forecasting of wildfire occurrence and spread is becoming increasingly important~\cite{pickell2017early, kotroni2020disarm}. Recent Earth foundation models (Earth FMs), pretrained on large-scale atmospheric and geophysical data~\cite{bodnar2025aurora, schmude2024prithviwxc, nguyen2023climax}, provide transferable representations for Earth-system dynamics and have shown strong performance across weather and remote-sensing tasks. However, wildfire dynamics depend on complex interactions among weather, vegetation, topography, fuel conditions, and active-fire behavior, which are not explicitly modeled during pretraining in existing general-purpose Earth FMs. This mismatch raises a natural question: can representations learned for general atmospheric or geophysical objectives transfer reliably to wildfire forecasting, and how should we measure that transfer?

Answering this question requires solving two intertwined problems. The first is that existing Earth FMs are not pretrained specifically for wildfire dynamics, but instead adapted to wildfire tasks after general-purpose pretraining. To address this limitation, we introduce \textbf{\ourfm}, the first foundation model pretrained specifically for wildfire prediction using fire-relevant multimodal data, including regional weather dynamics, active-fire observations, topography, vegetation, and static environmental context. By incorporating wildfire-specific signals directly during pretraining, \ourfm\ learns representations aligned with the physical processes underlying fire behavior rather than relying on transfer from general atmospheric or geophysical objectives.

The second problem is evaluation: even with a domain-specific model, reliably comparing \ourfm\ against transferred general-purpose Earth FMs remains difficult. Wildfire events are sparse in space and time~\cite{ebert2009neighborhood, gilleland2009intercomparison}, making transfer conclusions highly sensitive to three sources of evaluation variability.
\textit{First,} matching rules determine what counts as a correct prediction. Early warning systems tolerate spatial offsets that post-fire damage assessment cannot, so different matching rules can produce substantially different F1 scores from the same model outputs~\cite{ebert2009neighborhood, gilleland2009intercomparison}.
\textit{Second,} head-selection metrics determine which lightweight adapter is chosen on top of a frozen representation. Ranking metrics such as PR-AUC and decision metrics such as F1 can favor different heads from the same frozen features~\cite{mcdermott2024aurocauprc, traub2024selectiveclassification}.
\textit{Third,} wildfire task forms operate over different prediction units and metric families. Occupancy prediction, spread forecasting, burned-area regression, and analog retrieval therefore produce scores that are not directly comparable even under the same backbone~\cite{schaeffer2023mirage, gerard2023wildfirespreadts}.
Related protocol sensitivity has also been observed in active learning~\cite{luth2023activelearning} and selective classification~\cite{traub2024selectiveclassification}. We show that these effects become particularly severe in wildfire transfer evaluation, where sparse events and heterogeneous task forms amplify evaluation instability~\cite{marsocci2024pangaea}.

\begin{figure*}
    \centering
    \includegraphics[width=0.96\linewidth]{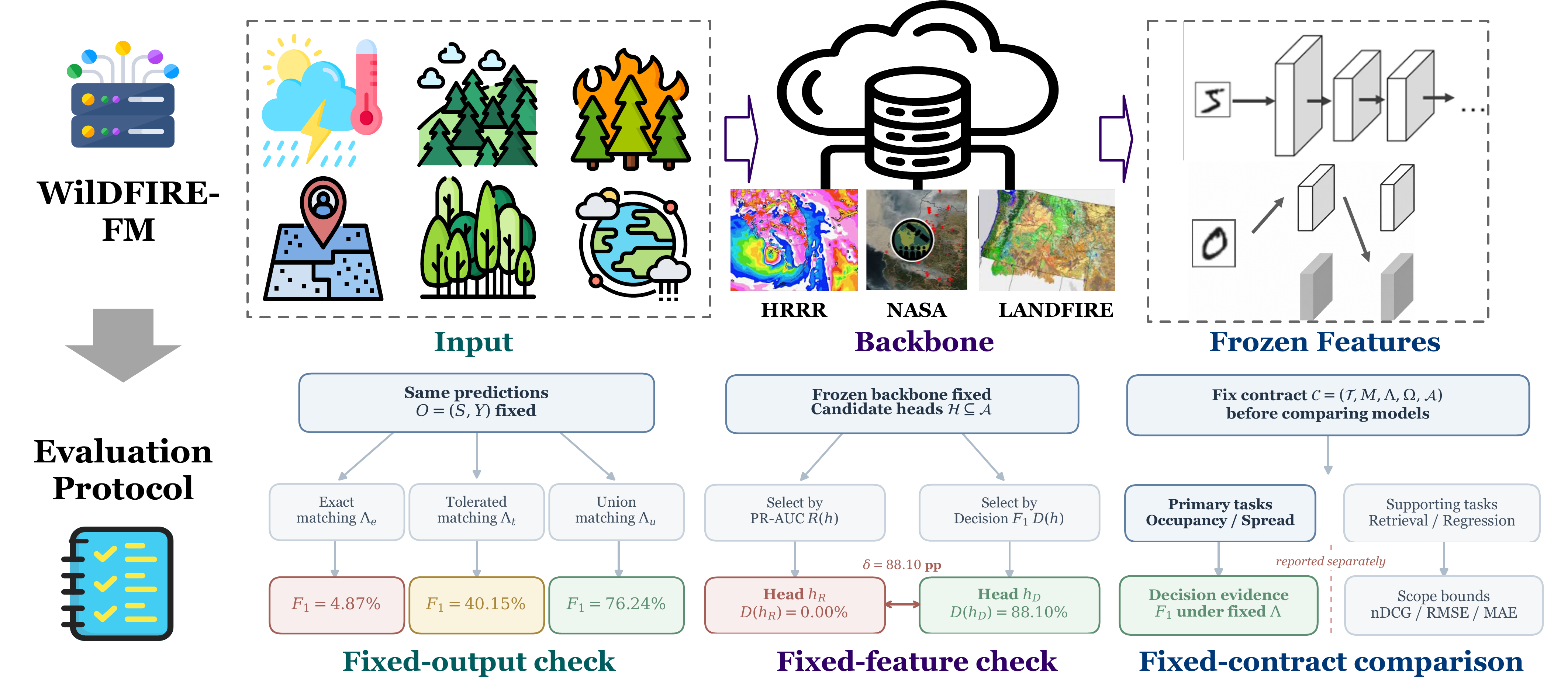}
    \caption{Overview of \textbf{\ourfm} and \textbf{Evaluation Protocol} in this paper.}
    \vspace{-8mm}
    \label{fig:overview}
\end{figure*}

This evaluation instability makes reliable comparison between 
\ourfm\ and existing Earth FMs fundamentally difficult. Standard 
geospatial benchmarks such as GEO-Bench~\cite{lacoste2023geobench}, 
WeatherBench2~\cite{rasp2024weatherbench2}, WILDS~\cite{koh2021wilds}, 
SustainBench~\cite{yeh2021sustainbench}, and 
TorchGeo~\cite{torchgeo2022} standardize datasets, splits, and 
metrics, but do so for tasks with dense, balanced labels where 
matching-rule sensitivity is not a primary concern. Wildfire 
studies such as FireCast~\cite{radke2019firecast} and Next Day 
Wildfire Spread~\cite{huot2022nextday} apply the same 
report-and-compare paradigm directly to sparse fire events 
without controlling for matching-rule choice, head-selection 
metric, or task-form comparability, the three sources of 
instability identified above. As a result, scores reported under 
different implicit protocol choices are not directly comparable, 
even when the underlying predictions are identical.

Based on the \textit{limitations of prior work}, our contributions are as follows (see Figure~\ref{fig:overview}).
\begin{itemize}

\item \textbf{Wildfire-specific foundation model.}
We introduce \ourfm, the first foundation model pretrained
specifically for wildfire prediction using multimodal wildfire
data spanning weather, active-fire observations, topography,
vegetation, and static environmental context. Unlike general
Earth FMs adapted after pretraining, \ourfm learns wildfire
representations directly from fire-relevant processes during
pretraining.

\item \textbf{Fixed-contract evaluation framework.}
We formulate wildfire Earth-FM transfer as a fixed-contract
evaluation problem, defining a contract
$\mathcal{C} = (\mathcal{T}, M, \Lambda, \Omega, \mathcal{A})$
that specifies the task, metric, matching rule, evaluation
scope, and lightweight-head family before comparison. We
introduce two controlled checks: a \emph{fixed-output check}
for matching-rule effects and a \emph{fixed-feature check}
for head-selection effects, enabling evaluation artifacts to
be separated from representation quality.

\item \textbf{Systematic wildfire transfer benchmark.}
Under fixed contracts, we compare \ourfm\ against ten
general-purpose Earth FMs across six wildfire task forms.
Our results show that wildfire transfer conclusions are
highly sensitive to evaluation design and strongly
task-dependent across occupancy, spread, retrieval, and
regression settings.

\end{itemize}


%% file: sections/2_backbone.tex
\section{\ourfm Reference Backbone}
\label{Reference_backbone}

\ourfm is a wildfire-specialized regional backbone trained on fire-relevant multimodal data for wildfire prediction. Existing general-purpose Earth FMs are pretrained for atmospheric and geophysical objectives~\cite{lam2023graphcast}, or for remote-sensing objectives~\cite{reed2023scalemae}, so wildfire-relevant information enters only indirectly through those objectives. In contrast, \ourfm is trained with weather, active-fire observations, topography, vegetation, and static environmental context, so its representation is learned from inputs tied directly to wildfire behavior. This design makes \ourfm a strong wildfire-specific backbone whose features are shaped by signals directly relevant to fire occurrence and spread.
It provides a task-aligned regional model trained directly for wildfire prediction.
It also serves as an empirical anchor for interpreting how transferred Earth FMs behave under matched evaluation contracts. This section describes the data resources and training strategy used to build \ourfm as an in-domain reference backbone. The fixed-contract protocol used to compare it with transferred Earth FMs is defined separately in Section~\ref{sec:eval}.

\subsection{Data Resources}
We group the resources by their role in the study: dynamic weather inputs, occupancy supervision, static context, and event-level resources for supporting tasks. Source and terms-of-use notes for the external data and model assets used in this study are summarized in Appendix Table~\ref{tab:external_assets_licenses}.

\noindent\textbf{Dynamic weather inputs.}
The weather inputs come from a California regional dataset built from NOAA High-Resolution Rapid Refresh (HRRR) fields~\cite{noaa_hrrr_ncei,noaa_hrrr_emc}. The data are placed on a projected 5 km grid in EPSG:5070. Each time map uses weather fields every 6 hours and predicts wildfire occupancy at a 12-hour lead. The variables include near-surface temperature and dew point, wind, CAPE, surface pressure, boundary-layer height, visibility, precipitation rate, and accumulated precipitation.

\noindent\textbf{Occupancy supervision.}
Wildfire supervision comes from NASA FIRMS active-fire detections~\cite{nasa_firms}. The detections are mapped to the same grid as the weather fields. \ourfm is trained on gridded occupancy labels derived from these detections. This defines the occupancy target used by the reference backbone throughout the primary experiments.

\noindent\textbf{Static context.}
Static context describes landscape and exposure factors that do not change at the weather time step. These variables are LANDFIRE fire-behavior fuel model~\cite{landfire_fbfm40}, LANDFIRE canopy cover~\cite{landfire_canopy_cover}, Wildfire Risk to Communities housing-unit density~\cite{usfs_wrc_housing_density}, and LandScan population~\cite{ornl_landscan_2024}. Together with validity masks for the weather and static fields, the occupancy input has 16 channels: 10 weather fields, two validity masks, and four static layers for regional fire prediction.

\noindent\textbf{Event-level resources.}
Event-level resources are used for supporting burned-area and analog tasks, not as occupancy labels for \ourfm. These resources include WFIGS incident and perimeter attributes~\cite{nifc_wfigs_perimeters} and MTBS burned-area and burn-severity records~\cite{mtbs_usgs_2025}. They provide event-scale outcomes and incident metadata for supporting tasks in the experiments and appendix analyses.

\subsection{Training Strategy}

\noindent\textbf{Model and data split.}
\ourfm uses a compact U-Net~\cite{ronneberger2015unet} that maps gridded weather and static inputs to wildfire predictions.
Its primary output is fire occupancy on the common spatial grid.
Data are split by time: June--August 2024 for training, September 2024 for validation, and October 2024 for testing.
This yields 368 training time maps, 120 validation time maps, and 120 test time maps.
Temporal splitting keeps later fire outcomes out of earlier training periods.

\noindent\textbf{Fire-aware tile training.}
Training is performed on 32$\times$32 tiles sampled from the time maps. The tiles include fire-centered regions and non-fire context, so the model sees both sparse fire labels and surrounding background conditions. This sampling reduces the dominance of empty cells without removing non-fire examples from the training distribution. Class-weighted binary cross-entropy is used for the primary occupancy target to further balance sparse positives.

\noindent\textbf{Spatial-support training objective.}
Wildfire labels can shift by a few grid cells because detections, weather fields, and static layers are aligned on a common grid. To reduce sensitivity to these small displacements during training, the occupancy target is dilated by two grid cells. An auxiliary spatial-support output is trained for the same neighborhood alongside the primary occupancy output. At test time, \ourfm is scored under the same task-specific evaluation contracts as the transferred Earth-FM backbones in Section~\ref{sec:eval}, ensuring matched comparison conditions.

%% file: sections/3_prelim.tex
\section{Evaluation Design}
\label{sec:eval}


\subsection{Wildfire Output Records and Fire Sets}

\begin{figure}
    \centering
        \vspace{-2mm}
    \includegraphics[width=0.8\linewidth]{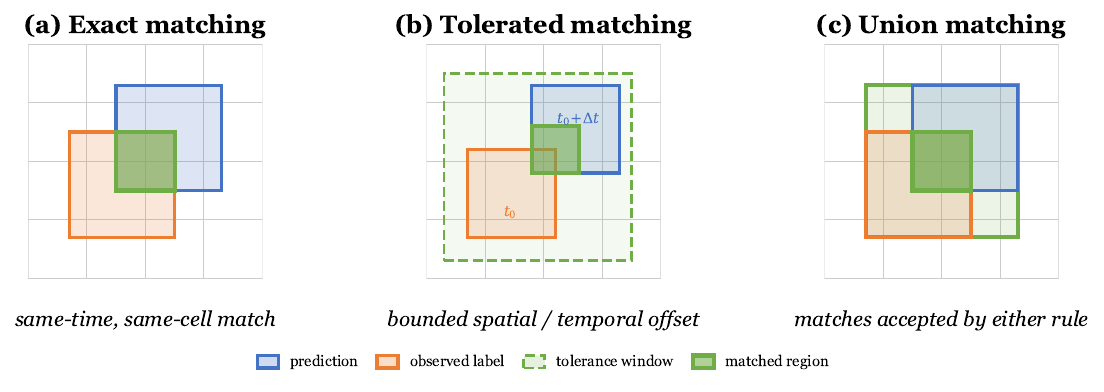}
    \vspace{-2mm}
    \caption{Matching rules for one fixed occupancy output.
    (a) Exact matching counts only same-time, same-cell overlap.
    (b) Tolerated matching accepts bounded spatial or temporal offsets.
    (c) The union reading counts matches accepted by either rule.}
    \vspace{-5mm}
    \label{fig:toy_occupancy_contract}
\end{figure}

\paragraph{Output record.}
A wildfire prediction model produces scores over spatial units and forecast times, which are compared against observed fire activity to compute performance. We formalize this comparison as a \emph{wildfire output record} $\mathcal{O} = (S, Y)$, where the score field $S = \{s_{i,t}\}$ contains model scores over spatial units $i$ and times $t$, and the label field $Y = \{y_{i,t}\}$ contains the corresponding observations. For occupancy tasks, $y_{i,t} \in \{0,1\}$ indicates whether fire is observed at $(i,t)$.
\vspace{-0.5em}
\paragraph{Predicted and observed fire sets.}
To evaluate $\mathcal{O}$, the score field is thresholded at $\tau$ to produce a predicted fire set $\hat{P}_\tau = \{(i,t) : s_{i,t} \geq \tau\}$, while the observed fire set is $P = \{(i,t) : y_{i,t} = 1\}$. The pair $(\hat{P}_\tau, P)$ is evaluated under a matching rule. Given a matching rule, true positives (TP), false positives (FP), and false negatives (FN) are computed from matched and unmatched elements, and the decision F1 score is $\text{F1} = 2\text{TP}/(2\text{TP} + \text{FP} + \text{FN})$. The same $(\hat{P}_\tau, P)$ can yield different TP, FP, and FN counts under different matching rules without changing model outputs, motivating the fixed-output check in Section~\ref{sec:checks}.
\vspace{-0.5em}
\paragraph{Matching rules.}
A matching rule specifies when a predicted unit-time pair in $\hat{P}_\tau$ is considered a match to an observed pair in $P$~\cite{ebert2009neighborhood, gilleland2009intercomparison}. Because wildfire applications tolerate different levels of spatial and temporal error, we define three matching rules for occupancy outputs. \textit{(1) Exact matching}: requires agreement in both spatial unit and forecast time. \textit{(2) Tolerated matching}: accepts predictions within a fixed spatial or temporal neighborhood defined by the evaluation contract $\mathcal{C}$. \textit{(3) Union matching}: accepts predictions satisfying either exact or tolerated matching.
%
Figure~\ref{fig:toy_occupancy_contract} illustrates these rules for a fixed output. Because the output record is held constant, any score difference is attributed solely to the matching rule.

\subsection{Evaluation Contract}

A wildfire transfer score depends not only on the model, but also on the evaluation choices used to compute it~\cite{luth2023activelearning}. Changing the matching rule $\Lambda$, metric $M$, or evaluation scope $\Omega$ changes what the score measures even when model outputs are fixed.

We define an \emph{evaluation contract} as the tuple
$\mathcal{C} = (\mathcal{T}, M, \Lambda, \Omega, \mathcal{A})$,
where $\mathcal{T}$ denotes the task, $M$ the metric,
$\Lambda$ the matching rule, $\Omega$ the evaluation scope,
and $\mathcal{A}$ the allowed lightweight-head family.
Two transfer scores are comparable only when all five
components are identical. The evaluation scope $\Omega$ is particularly important in wildfire settings. A global scope evaluates the full spatial domain, including many fire-inactive regions that can mask differences between models. A fire-prone scope restricts evaluation to regions with higher historical fire activity. We report both scopes separately rather than averaging across them. Fixed matching-rule, task-form, and scope parameters are reported in Appendix Tables~\ref{tab:app_matching_rule_params}, \ref{tab:app_contract_params_full}, and~\ref{tab:app_scope_params}.

\subsection{Task-Form Contracts}
\label{sec:taskforms}
Contract components depend on task form. We distinguish \emph{primary} and \emph{supporting} tasks based on whether they directly evaluate wildfire decisions. Occupancy and fire spread are primary tasks because they evaluate spatial fire outputs under matching or overlap rules.
Retrieval, burned-area regression, smoke PM$_{2.5}$, and extreme heat are supporting tasks because they use different prediction units and metric families. Their results provide complementary evidence rather than direct substitutes for occupancy and spread evaluation~\cite{schaeffer2023mirage}.

\begin{wrapfigure}[19]{r}{0.52\textwidth}
\vspace{-2em}
\centering
\includegraphics[width=\linewidth]{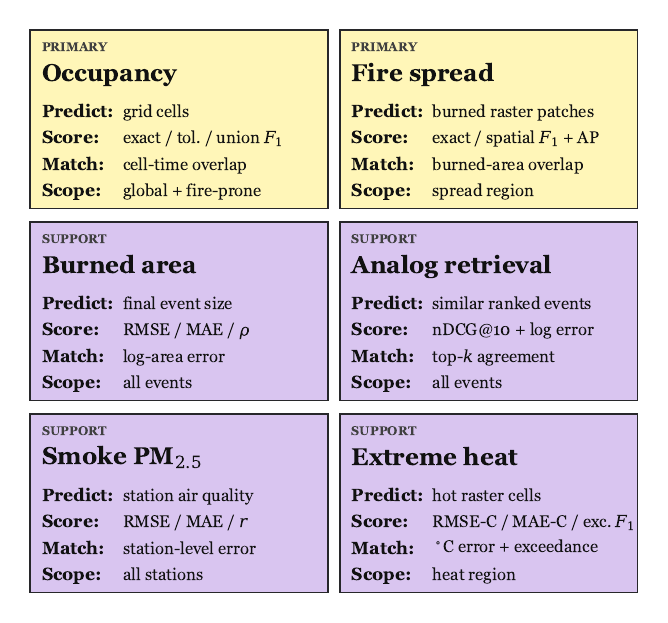}
\vspace{-1.5em}
\caption{
Evaluation contract map for the six fixed-contract tasks.
Yellow boxes denote \textcolor[RGB]{255,193,7}{\textbf{primary}} decision tasks; purple boxes denote \textcolor[RGB]{148,103,189}{\textbf{supporting}} tasks.
}
\label{fig:task_contract_tiles}
\vspace{-0.8em}
\end{wrapfigure}

For primary tasks, multiple metrics are reported for the same output under different contracts. For occupancy, exact F1 requires same-cell same-time agreement, tolerated F1 accepts predictions within a spatial or temporal neighborhood, and union F1 accepts predictions satisfying either rule. For fire spread, exact F1 evaluates raster-cell agreement, spatial F1 evaluates region overlap between $\hat{B}$ and $B$~\cite{gilleland2009intercomparison}, and AP summarizes ranking quality across thresholds. These metrics are reported separately rather than aggregated because they measure different aspects of the same prediction task. Figure~\ref{fig:task_contract_tiles} summarizes the contract map across all six task forms.

\subsection{Controlled Checks}
\label{sec:checks}
We isolate the two instability sources with two checks.
Each check fixes all contract components except one, so any difference is attributed solely to that component.

\paragraph{Fixed-output check.}
The fixed-output check isolates matching-rule effects by holding the output record $\mathcal{O} = (S, Y)$ and all other contract components fixed while varying only $\Lambda$. For the same occupancy record, we compute F1 under exact, tolerated, and union matching. Any score difference is therefore attributed solely to the matching rule. If matching rules alone shift F1 by tens of percentage points on the same output, then comparing models under different $\Lambda$ conflates model quality with evaluation design.

\paragraph{Fixed-feature check and selection regret.}
The fixed-feature check isolates head-selection effects by holding the frozen feature source, $\mathcal{T}$, $\Omega$, $\Lambda$, and candidate head family $\mathcal{H} \subseteq \mathcal{A}$ fixed while varying only the selection metric. Let $R(h)$ denote the ranking score of head $h$ and $D(h)$ its decision score. Ranking-based selection chooses $h_R = \arg\max_{h \in \mathcal{H}} R(h)$, while decision-based selection chooses $h_D = \arg\max_{h \in \mathcal{H}} D(h)$. We define \emph{selection regret} as the decision-score gap incurred by using a ranking metric as a proxy for a decision metric during head selection: $\delta = D(h_D) - D(h_R) \geq 0$ ~\cite{mcdermott2024aurocauprc, traub2024selectiveclassification}. When $\delta > 0$, the ranking metric selects a head with lower decision performance under the same frozen representation, indicating that the observed gap arises from metric misalignment rather than from representation quality.The head family used in fixed-feature comparisons is summarized in Appendix Table~\ref{tab:app_head_architectures}.

\paragraph{Fixed-contract transfer comparison.}
After the controlled checks establish that matching-rule and selection-metric effects are non-trivial, Earth-FM backbones are evaluated under a shared contract $\mathcal{C}$. Entries are compared only when they satisfy the same $(\mathcal{T}, M, \Lambda, \Omega, \mathcal{A})$ tuple. Supporting tasks test whether occupancy and spread patterns generalize across task forms and provide additional evidence when transfer orderings are preserved.

%% file: sections/4_experiments.tex
\section{Experiments}
\label{sec:experiments}

We address three research questions under the fixed-contract framework defined in Section~\ref{sec:eval}. \textbf{RQ1:} Under fixed outputs, does the matching rule determine whether a wildfire model appears usable?
\textbf{RQ2:} Under fixed features, does ranking-based head selection lose decision performance?
\textbf{RQ3:} Under fixed task contracts, do model comparisons remain consistent across task forms?
\vspace{-0.5em}
\subsection{Experimental Setup}
\paragraph{Task instances.}
We instantiate the six task-form contracts defined in Section~\ref{sec:taskforms}.
Occupancy and fire spread serve as primary tasks because they evaluate spatial fire outputs under matching or overlap rules and align with the decision structure of early warning systems~\cite{goldammer1999early, farahmand2020fdeo}.
The four supporting tasks, \textit{final burned area, analog retrieval, smoke PM$_{2.5}$, and extreme heat}, use different prediction units and metric families; their results bound rather than replace primary decision evidence.

\paragraph{Compared backbones.}
The frozen Earth-FM comparator set includes Prithvi-WxC~\cite{schmude2024prithviwxc}, Aurora~\cite{bodnar2025aurora}, ClimaX~\cite{nguyen2023climax}, StormCast~\cite{pathak2024stormcast}, DLWP~\cite{weyn2020dlwp}, FCN~\cite{pathak2022fourcastnet}, FengWu~\cite{chen2023fengwu}, FuXi~\cite{chen2023fuxi}, Pangu-Weather~\cite{bi2023panguweather}, and AlphaEarth~\cite{brown2025alphaearth}.
\ourfm\ serves as the wildfire-specialized reference backbone.

\paragraph{Protocol.}
For each comparison, the contract $\mathcal{C} = (\mathcal{T}, M, \Lambda, \Omega, \mathcal{A})$ is fixed before reporting test scores. 
Thresholds and morphology parameters are selected on validation data and held fixed at test time.
Stochastic components are evaluated over five seeds and reported as mean $\pm$ standard deviation; deterministic fixed-output checks have zero seed variance by construction.
Entries outside a fixed contract are omitted from main tables and documented in the appendix.
For error metrics lower is better ($\downarrow$); for F1, AP, nDCG, and correlation metrics higher is better ($\uparrow$).
Appendix Table~\ref{tab:app_seed_robustness} summarizes the seed-level checks behind the reported mean-with-std convention.

\subsection{Matching-Rule Sensitivity Under Fixed Output (RQ1)}
\label{sec:rq1}

To answer RQ1, we conduct a fixed-output check on occupancy and fire spread tasks, holding the score field $S$, label field $Y$, threshold, and all other operating choices fixed while varying only the matching rule $\Lambda$ across exact, tolerated, and union settings. Occupancy results are reported in Figure~\ref{fig:fireprone_contract_progression} under both global and fire-prone scopes. The same progression is applied to fire spread outputs. Complete occupancy sweeps and predicted-positive rates are reported in Appendix Tables~\ref{tab:fireprone_contract_progression} and~\ref{tab:app_occupancy_ppr_scope}.

\begin{wrapfigure}[21]{r}{0.50\textwidth}
\centering
\vspace{-3mm}
\includegraphics[width=\linewidth]{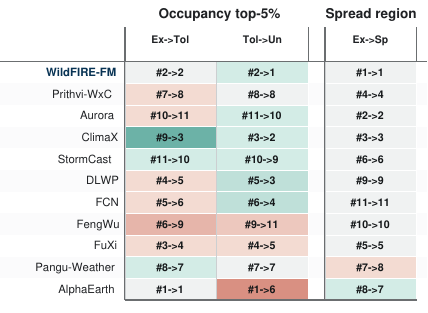}
\caption{\textbf{Primary-task rank changes (RQ1).}
Cells show rank before\(\rightarrow\)after. Green/red/gray mark moving up/down/no change; darker green or red marks a larger move. Following Section~\ref{sec:taskforms}, Ex/Tol/Un are occupancy exact, tolerated, and union matching; Sp is spread spatial-overlap $F_1$.}
\label{fig:primary_ranking}
\vspace{-0.8em}
\end{wrapfigure}
Because both tasks involve spatially sparse targets, fire-active cells for occupancy, burned raster patches for spread, the operational assumptions encoded in $\Lambda$ directly govern what the model is being asked to get right, making matching-rule choice a substantive experimental setting rather than a post hoc evaluation detail.
The fixed-output results reveal a pattern that goes beyond score differences: matching-rule choice determines whether a model appears viable for wildfire decision tasks at all. Under exact matching, which requires same-cell same-time agreement, the majority of frozen Earth-FM backbones produce F1 scores that are effectively near zero, rendering them indistinguishable from an uninformative baseline and suggesting they have no practical utility for the task. As the matching rule relaxes to tolerated and then union matching, both of which reflect operationally realistic assumptions for early warning systems, where a prediction displaced by a few grid cells still triggers the correct response, the same frozen representations recover substantial decision performance, with several backbones crossing from near-zero to practically meaningful F1 levels. This transition is not a marginal score improvement: it is a qualitative change in whether a model can be considered usable. The same pattern holds for fire spread under region-level matching relaxation, where strict raster-cell agreement again suppresses performance for most backbones while spatial tolerance restores it. The implications for prior wildfire transfer claims are significant: papers that report model performance under a single implicit matching rule, which is common practice given that sparse decision targets almost always require some form of tolerance~\cite{ebert2009neighborhood, gilleland2009intercomparison}, may be drawing viability conclusions that are entirely dependent on an undisclosed protocol choice. A model claimed to perform well under one tolerance assumption may be completely unusable under a stricter one, and vice versa. Matching rule cannot be treated as an evaluation detail; it is an experimental setting that must be fixed, reported, and justified as part of any wildfire transfer claim. Additional spread AP values under fixed scopes are reported in Appendix Table~\ref{tab:app_spread_ap_by_scope}.

\begin{table}[t]
\centering
\small
\setlength{\tabcolsep}{4pt}
\renewcommand{\arraystretch}{1.20}
\caption{%
  \textbf{Primary fixed-contract transfer results (RQ1).}
  Occupancy metrics: exact, tolerated, union $F_1$ (\%).
  Fire spread metrics: exact $F_1$ and spatial $F_1$ (\%).
  Each block fixes $\mathcal{T}$, $\Lambda$, $\Omega$, and $\mathcal{A}$.
  Upward arrows indicate that larger values are better.
  \textbf{Bold} marks the best value per metric. \textbf{Tol.} = Tolerated
}
\label{tab:primary_results}
\setlength{\arrayrulewidth}{0.4pt}
\resizebox{\textwidth}{!}{%
\begin{tabular}{lccccc}
\toprule
& \multicolumn{3}{c}{\textbf{Occupancy}} 
& \multicolumn{2}{c}{\textbf{Fire spread}} \\
\cmidrule(lr){2-4}\cmidrule(lr){5-6}
\textbf{Comparator} 
& \textbf{Exact $F_1\uparrow$} & \textbf{Tol.\ $F_1\uparrow$} & \textbf{Union $F_1\uparrow$}
& \textbf{Exact $F_1\uparrow$} & \textbf{Spatial $F_1\uparrow$} \\
\midrule
\ourfm\
& \ms{0.4546}{0.1412}
& \ms{29.7484}{1.2868}
& \ms{59.0656}{2.7372}
& \ensuremath{\mathbf{37.6700}{\mkern1mu}_{\scriptscriptstyle \boldsymbol{\pm}\mathbf{0.9800}}}
& \ensuremath{\mathbf{80.9700}{\mkern1mu}_{\scriptscriptstyle \boldsymbol{\pm}\mathbf{2.0200}}} \\
\midrule
Prithvi-WxC
& \ms{0.0552}{0.0039} & \ms{7.1649}{0.6557} & \ms{20.1853}{1.8299}
& \ms{22.3500}{3.4500} & \ms{65.2600}{1.0700} \\
Aurora
& \ms{0.0656}{0.0094} & \ms{8.5009}{1.9594} & \ms{23.1037}{4.9418}
& \ms{30.8757}{0.1343} & \ms{71.7329}{0.0141} \\
ClimaX
& \ms{0.3480}{0.0754}
& \ensuremath{\mathbf{29.7535}{\mkern1mu}_{\scriptscriptstyle \boldsymbol{\pm}\mathbf{3.6073}}}
& \ensuremath{\mathbf{60.1506}{\mkern1mu}_{\scriptscriptstyle \boldsymbol{\pm}\mathbf{7.5865}}}
& \ms{27.9853}{2.0532} & \ms{69.0634}{2.3832} \\
StormCast
& \ms{0.0626}{0.0119} & \ms{8.1951}{2.1895} & \ms{22.3817}{5.4294}
& \ms{14.8387}{7.5791} & \ms{55.7568}{21.3003} \\
DLWP
& \ms{0.1693}{0.0419} & \ms{14.9148}{3.2446} & \ms{28.1901}{6.9658}
& \ms{5.9335}{10.0712} & \ms{22.8587}{22.3750} \\
FCN
& \ms{0.2829}{0.0839} & \ms{19.5061}{3.3412} & \ms{40.0604}{9.3701}
& \ms{3.1798}{2.6598} & \ms{15.6203}{12.4531} \\
FengWu
& \ms{0.2613}{0.0757} & \ms{12.0050}{6.0239} & \ms{24.1022}{13.6293}
& \ms{5.5189}{9.0883} & \ms{18.4774}{22.4703} \\
FuXi
& \ms{0.3774}{0.1212} & \ms{21.0323}{4.8211} & \ms{37.2888}{9.4470}
& \ms{19.9909}{2.1364} & \ms{56.1826}{3.0412} \\
Pangu-Weather
& \ms{0.2755}{0.1089} & \ms{17.0909}{4.0477} & \ms{35.6386}{9.0327}
& \ms{11.2583}{11.0719} & \ms{32.5081}{25.4969} \\
AlphaEarth
& \ensuremath{\mathbf{2.0606}{\mkern1mu}_{\scriptscriptstyle \boldsymbol{\pm}\mathbf{0.4404}}}
& \ms{29.4476}{6.0064} & \ms{37.4286}{9.9458}
& \ms{11.0995}{3.6088} & \ms{32.8316}{7.4634} \\
\bottomrule
\end{tabular}
}
\end{table}

\begin{wrapfigure}[14]{r}{0.50\textwidth}
    \centering
    \vspace{-1em} \includegraphics[width=0.50\textwidth]{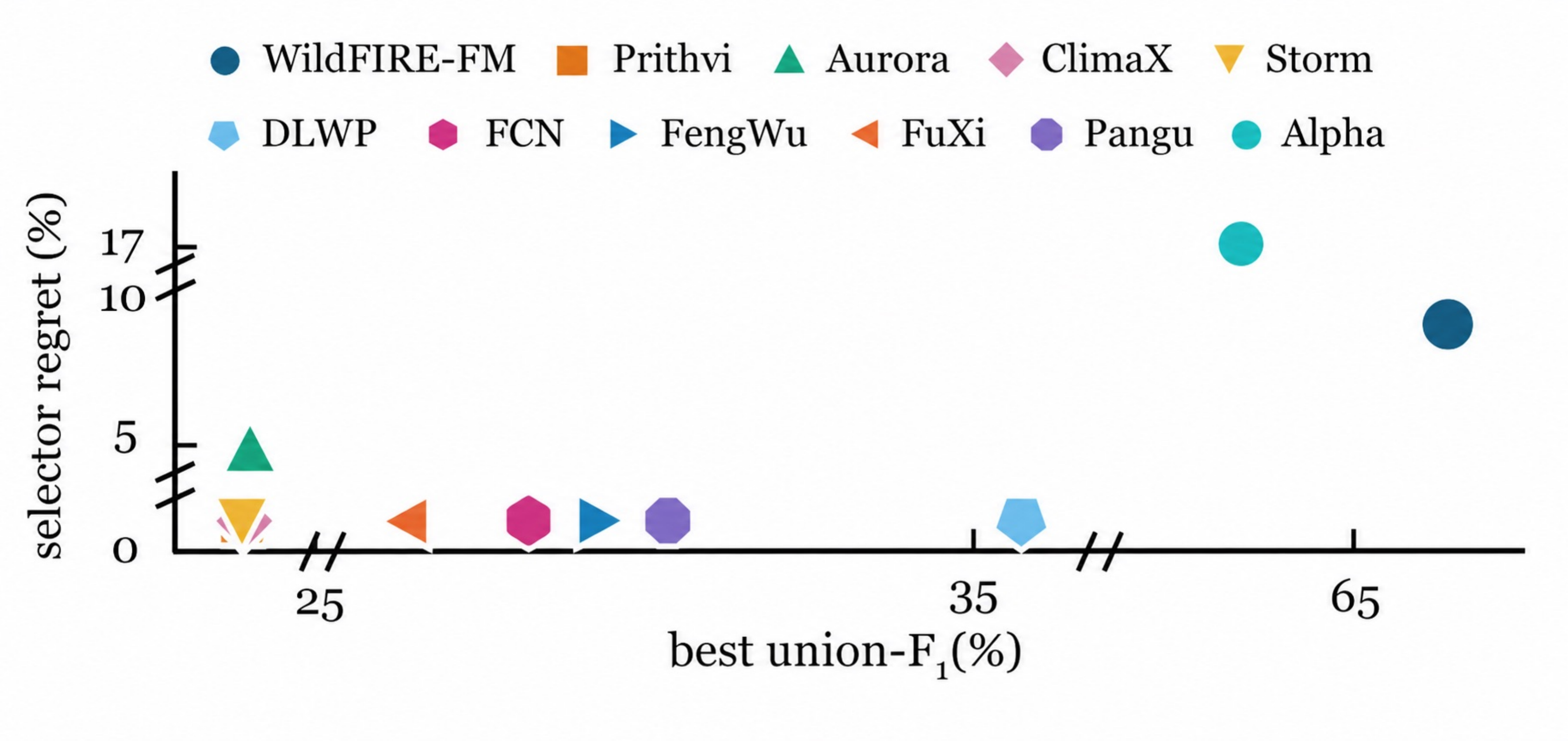}
\caption{\textbf{Head-selection regret under fixed features (RQ2).}
Each point is one backbone; selection regret \(\delta\) follows Section~\ref{sec:checks} under global-scope union-\(F_1\).}
    
    \label{fig:selection_regret_diagnostic}
    \vspace{-1.2em}
\end{wrapfigure}

\subsection{Head-Selection Sensitivity Under Fixed Features (RQ2)}
\label{sec:rq2}

To answer RQ2, we conduct a fixed-feature check on occupancy and fire spread tasks, holding the frozen feature source, $\mathcal{T}$, $\Omega$, $\Lambda$, and candidate head family $\mathcal{H} \subseteq \mathcal{A}$ fixed while varying only the selection metric between PR-AUC-based and decision-F1-based selection. The resulting selection regret $\delta = D(h_D) - D(h_R)$ measures the decision-score loss induced by metric misalignment. Occupancy results are reported in Figure~\ref{fig:selection_regret_diagnostic} under both global and fire-prone scopes. Full per-seed and per-head details are reported in Appendix~\ref{sec:app_seeded_audits}, and the exact, tolerated, and union regret breakdown is provided in Appendix Table~\ref{tab:appendix_selection_regret_tolerance}.

The fixed-feature results show that head-selection metrics introduce substantial backbone-dependent variation that is not explained by representation quality alone. Some backbones exhibit near-zero regret, indicating agreement between PR-AUC and decision-F1 selection, while others show large regret concentrated in specific scope-matching settings. Regret is generally larger under the global scope, where severe fire imbalance amplifies misalignment between ranking and decision metrics~\cite{mcdermott2024aurocauprc}. Restricting evaluation to fire-prone scopes typically reduces regret by concentrating evaluation on fire-relevant regions. A similar pattern appears for fire spread, where ranking and decision metrics can favor different heads under the same frozen representation. These results show that selection metrics must be aligned with the evaluation objective as part of the evaluation contract~\cite{traub2024selectiveclassification}.

\subsection{Supporting Task Checks (RQ3)}
\label{sec:rq3}

\begin{figure}[t]
    \centering
\vspace{-5mm}

    \includegraphics[width=\textwidth]{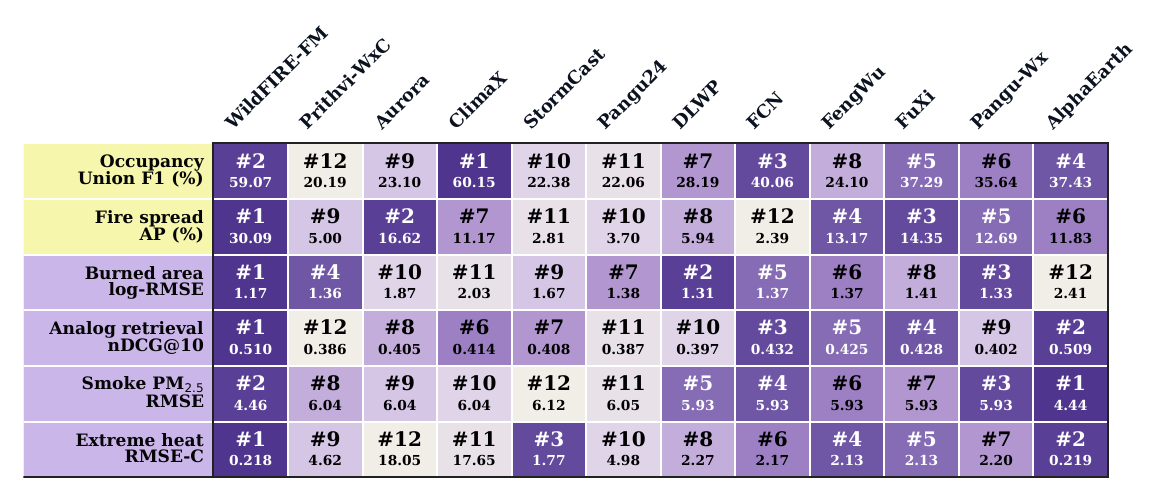}
    \vspace{-2mm}
    \caption{{\textbf{Rank map for supporting task comparison (RQ4).} Each row fixes one task contract $\mathcal{C}$ and ranks the eligible backbones within that contract. The figure shows rank changes across task forms; native metric values are reported in Table~\ref{tab:supporting_results}.}}
\vspace{-6mm}
    \label{fig:task_comparator_normalized_map}
\end{figure}

To answer RQ3, we evaluate all backbones across the four supporting task contracts, \textit{burned area, analog retrieval, smoke PM$_{2.5}$, }and\textit{ extreme heat}, and examine whether the reference-versus-frozen ordering established under primary tasks generalizes across task forms. A rank overview across all six contracts is provided in Figure~\ref{fig:task_comparator_normalized_map}, which maps backbone-by-task rank positions and makes cross-task ordering shifts directly visible. Native metric values are reported in Table~\ref{tab:supporting_results}. Additional supporting-task diagnostics are reported in Appendix Tables~\ref{tab:app_burned_area_median_acre}, \ref{tab:app_analog_rank_depth}, \ref{tab:app_smoke_high_event}, and~\ref{tab:app_heat_event_pr}.

The supporting task results produce three qualitatively distinct patterns relative to the primary findings. Burned area largely preserves the reference-versus-frozen ordering seen under occupancy and spread: \ourfm\ leads frozen entries on log-RMSE and Spearman $\rho$, suggesting that the representational advantage of wildfire-specific pretraining generalizes to event-scale regression under a different metric family, providing convergent evidence for the primary claim. Analog retrieval and smoke PM$_{2.5}$ show a different pattern, with AlphaEarth matching \ourfm\ closely on both tasks while atmospheric FMs show near-zero correlation on smoke PM$_{2.5}$, indicating that retrieval and air-quality signals are captured comparably by a general remote-sensing backbone, and that the primary occupancy advantage does not extend uniformly to these task forms. Extreme heat exhibits the largest variance across the comparator set, with atmospheric FMs ranging from near-reference performance to near-complete failure depending on backbone pretraining domain, while AlphaEarth again matches \ourfm\ closely. The scale of this variance is itself informative: aggregating scores across task forms without respecting contract boundaries would produce rankings dominated by scale artifacts in the extreme heat block rather than by transfer quality. Taken together, these results establish that supporting tasks bound rather than extend the primary claim, they provide useful evidence about where backbone families generalize and where they do not, but they cannot substitute for primary decision task evaluation, and their results must within their own task-form contracts.

\begin{table}[t]
\centering
\small
\setlength{\tabcolsep}{3.5pt}
\renewcommand{\arraystretch}{1.18}
\caption{%
  \textbf{Supporting task-metric matrix (RQ3).}
  Top: final burned area and analog retrieval.
  Bottom: smoke PM$_{2.5}$ and extreme heat.
  Each block fixes $\mathcal{T}$, $\Lambda$, and $\Omega$; backbone
  column is shared across paired tasks. \ourfm\ row is
  separated by a rule as the empirical anchor. \textbf{Bold} marks
  the best value per metric. For error metrics
  lower is better ($\downarrow$); for $F_1$, nDCG, and $r$ higher
  is better ($\uparrow$).
}
\label{tab:supporting_results}
\resizebox{\textwidth}{!}{%
\begin{tabular}{lcccccc}
\toprule
& \multicolumn{3}{c}{\textbf{Burned area}}
& \multicolumn{3}{c}{\textbf{Analog retrieval}} \\
\cmidrule(lr){2-4}\cmidrule(lr){5-7}
\textbf{Backbone}
& \textbf{log-RMSE$\downarrow$} & \textbf{log-MAE$\downarrow$}
& \textbf{Spearman$\uparrow$}
& \textbf{nDCG@10$\uparrow$} & \textbf{log-RMSE$\downarrow$}
& \textbf{log-MAE$\downarrow$} \\
\midrule
\ourfm\
& \ensuremath{\mathbf{1.1657}{\mkern1mu}_{\scriptscriptstyle \boldsymbol{\pm}\mathbf{0.0126}}}
& \ensuremath{\mathbf{1.0423}{\mkern1mu}_{\scriptscriptstyle \boldsymbol{\pm}\mathbf{0.0081}}}
& \ensuremath{\mathbf{0.6298}{\mkern1mu}_{\scriptscriptstyle \boldsymbol{\pm}\mathbf{0.0338}}}
& \ensuremath{\mathbf{0.5099}{\mkern1mu}_{\scriptscriptstyle \boldsymbol{\pm}\mathbf{0.0336}}}
& \ensuremath{\mathbf{1.1977}{\mkern1mu}_{\scriptscriptstyle \boldsymbol{\pm}\mathbf{0.1029}}}
& \ensuremath{\mathbf{1.0043}{\mkern1mu}_{\scriptscriptstyle \boldsymbol{\pm}\mathbf{0.0759}}} \\
\midrule
Prithvi-WxC
& \ms{1.3630}{0.0681} & \ms{1.2435}{0.0668} & \ms{0.1799}{0.3002}
& \ms{0.3857}{0.0189} & \ms{1.3908}{0.0938} & \ms{1.2585}{0.0865} \\
Aurora
& \ms{1.8658}{0.2009} & \ms{1.6717}{0.1245} & \ms{-0.1156}{0.2982}
& \ms{0.4046}{0.0144} & \ms{1.3659}{0.0792} & \ms{1.2596}{0.0968} \\
ClimaX
& \ms{2.0300}{0.2103} & \ms{1.8443}{0.1528} & \ms{-0.2515}{0.2688}
& \ms{0.4143}{0.0191} & \ms{1.4526}{0.0926} & \ms{1.2441}{0.1446} \\
StormCast
& \ms{1.6679}{0.1438} & \ms{1.4745}{0.1134} & \ms{0.1830}{0.1969}
& \ms{0.4076}{0.0094} & \ms{1.3663}{0.0781} & \ms{1.2371}{0.1078} \\
DLWP
& \ms{1.3070}{0.0980} & \ms{1.1769}{0.0834} & \ms{0.4888}{0.1368}
& \ms{0.3972}{0.0146} & \ms{1.5351}{0.0802} & \ms{1.3196}{0.0781} \\
FCN
& \ms{1.3693}{0.0885} & \ms{1.2599}{0.0723} & \ms{0.3484}{0.1662}
& \ms{0.4316}{0.0134} & \ms{1.4604}{0.1035} & \ms{1.2351}{0.0586} \\
FengWu
& \ms{1.3715}{0.1011} & \ms{1.2604}{0.0820} & \ms{0.3221}{0.2004}
& \ms{0.4246}{0.0237} & \ms{1.4179}{0.0986} & \ms{1.2233}{0.0915} \\
FuXi
& \ms{1.4068}{0.1011} & \ms{1.3023}{0.0789} & \ms{0.2663}{0.2561}
& \ms{0.4279}{0.0212} & \ms{1.4290}{0.0929} & \ms{1.2236}{0.0961} \\
Pangu-Weather
& \ms{1.3280}{0.0735} & \ms{1.2081}{0.0607} & \ms{0.4141}{0.1573}
& \ms{0.4017}{0.0245} & \ms{1.4235}{0.0731} & \ms{1.2225}{0.0847} \\
AlphaEarth
& \ms{2.4068}{0.2841} & \ms{2.0822}{0.2371} & \ms{-0.3428}{0.1716}
& \ms{0.5086}{0.0440} & \ms{1.2158}{0.1310} & \ms{1.0350}{0.1018} \\
\bottomrule
\end{tabular}
}

\vspace{4pt}

\resizebox{\textwidth}{!}{%
\begin{tabular}{lcccccc}
\toprule
& \multicolumn{3}{c}{\textbf{Smoke PM$_{2.5}$}}
& \multicolumn{3}{c}{\textbf{Extreme heat}} \\
\cmidrule(lr){2-4}\cmidrule(lr){5-7}
\textbf{Backbone}
& \textbf{RMSE$\downarrow$} & \textbf{MAE$\downarrow$}
& \textbf{Pearson $r\uparrow$}
& \textbf{RMSE-C$\downarrow$} & \textbf{MAE-C$\downarrow$}
& \textbf{Exceed.\ $F_1\uparrow$} \\
\midrule
\ourfm\
& \ms{4.4646}{0.0060}
& \ms{2.4108}{0.0016}
& \ensuremath{\mathbf{0.6368}{\mkern1mu}_{\scriptscriptstyle \boldsymbol{\pm}\mathbf{0.0013}}}
& \ensuremath{\mathbf{0.2179}{\mkern1mu}_{\scriptscriptstyle \boldsymbol{\pm}\mathbf{0.0043}}}
& \ensuremath{\mathbf{0.1787}{\mkern1mu}_{\scriptscriptstyle \boldsymbol{\pm}\mathbf{0.0018}}}
& \ms{0.9541}{0.0164} \\
\midrule
Prithvi-WxC
& \ms{6.0382}{0.0828} & \ms{3.7301}{0.0055} & \ms{0.0243}{0.0045}
& \ms{4.6225}{0.0192} & \ms{2.6315}{0.0128} & \ms{0.8693}{0.0023} \\
Aurora
& \ms{6.0384}{0.0828} & \ms{3.7265}{0.0055} & \ms{0.0193}{0.0043}
& \ms{18.0474}{0.0708} & \ms{15.3747}{0.0594} & \ms{0.0951}{0.0038} \\
ClimaX
& \ms{6.0402}{0.0828} & \ms{3.7290}{0.0055} & \ms{0.0004}{0.0029}
& \ms{17.6492}{0.0347} & \ms{14.4938}{0.0319} & \ms{0.7684}{0.0068} \\
StormCast
& \ms{6.1230}{0.0830} & \ms{3.8182}{0.0073} & \ms{0.0183}{0.0041}
& \ms{1.7671}{0.2145} & \ms{1.3507}{0.1576} & \ms{0.9073}{0.0189} \\
DLWP
& \ms{5.9289}{0.1031} & \ms{3.7331}{0.0088} & \ms{0.0303}{0.0060}
& \ms{2.2662}{0.1106} & \ms{1.7153}{0.0748} & \ms{0.9156}{0.0112} \\
FCN
& \ms{5.9277}{0.1033} & \ms{3.7345}{0.0088} & \ms{0.0312}{0.0062}
& \ms{2.1657}{0.1800} & \ms{1.6033}{0.1039} & \ms{0.9257}{0.0096} \\
FengWu
& \ms{5.9297}{0.1032} & \ms{3.7395}{0.0088} & \ms{0.0304}{0.0063}
& \ms{2.1266}{0.1589} & \ms{1.5801}{0.1004} & \ms{0.0481}{0.0459} \\
FuXi
& \ms{5.9319}{0.1029} & \ms{3.7398}{0.0088} & \ms{0.0299}{0.0061}
& \ms{2.1282}{0.0969} & \ms{1.5759}{0.0719} & \ms{0.2268}{0.0623} \\
Pangu-Weather
& \ms{5.9270}{0.1036} & \ms{3.7320}{0.0088} & \ms{0.0301}{0.0060}
& \ms{2.2045}{0.1483} & \ms{1.6307}{0.0889} & \ms{0.0199}{0.0062} \\
AlphaEarth
& \ensuremath{\mathbf{4.4403}{\mkern1mu}_{\scriptscriptstyle \boldsymbol{\pm}\mathbf{0.0488}}}
& \ensuremath{\mathbf{2.3992}{\mkern1mu}_{\scriptscriptstyle \boldsymbol{\pm}\mathbf{0.0056}}}
& \ms{0.6347}{0.0066}
& \ms{0.2194}{0.0039}
& \ms{0.1800}{0.0014}
& \ensuremath{\mathbf{0.9542}{\mkern1mu}_{\scriptscriptstyle \boldsymbol{\pm}\mathbf{0.0107}}} \\
\bottomrule
\end{tabular}
}
\end{table}

\paragraph{Pattern 1: primary pattern preserved (burned area).}
\ourfm\ leads all frozen entries on log-RMSE and Spearman $\rho$. The ordering observed under occupancy and spread is preserved under burned-area regression despite the different prediction unit and metric family.

\paragraph{Pattern 2: primary pattern bounded (analog retrieval and smoke PM$_{2.5}$).}
For analog retrieval, AlphaEarth matches \ourfm\ (nDCG@10 $= 0.51 \pm 0.04$ vs.\ $0.51 \pm 0.03$). For smoke PM$_{2.5}$, AlphaEarth also matches \ourfm\ on MAE and Pearson $r$, while atmospheric Earth FMs show near-zero correlation. These results show that the occupancy-and-spread ordering does not fully extend to all supporting tasks once AlphaEarth is included.

\paragraph{Pattern 3: primary pattern bounded with large variance (extreme heat).} AlphaEarth matches \ourfm on RMSE-C and remains close on exceedance F1, while atmospheric FMs range from RMSE-C $= 1.77$ (StormCast) to $18.05$ (Aurora). This large spread indicates that aggregated scores across task forms would be dominated by scale artifacts rather than transfer quality, reinforcing the need for per-contract reporting established in Section~\ref{sec:eval}.

\textit{Answer to RQ3:} Figure~\ref{fig:task_comparator_normalized_map} and Table~\ref{tab:supporting_results} show that burned area preserves the primary reference-versus-frozen pattern under a different metric family. Analog retrieval, smoke PM$_{2.5}$, and extreme heat bound this pattern: AlphaEarth matches or approaches \ourfm on these tasks, indicating that the primary occupancy and spread claims do not extend uniformly across all task forms.

%% file: sections/5_conclusion.tex
\section{Conclusion}

We introduced \ourfm, the first foundation model pretrained
specifically for wildfire prediction using fire-relevant
multimodal data. Our results show that wildfire forecasting
requires representations aligned with wildfire dynamics rather
than transfer alone from general atmospheric or geophysical
pretraining.
At the same time, our study shows that reliable wildfire
transfer evaluation is substantially more difficult than
standard benchmark settings suggest. Wildfire transfer
conclusions depend strongly on matching rules, head-selection
metrics, and task form, and scores computed under different
evaluation settings are not directly comparable. These effects
become particularly pronounced in sparse spatiotemporal
prediction settings such as wildfire forecasting.
We therefore introduced a fixed-contract evaluation framework
for wildfire Earth-FM transfer. By explicitly specifying the
task, metric, matching rule, evaluation scope, and head family
before comparison, fixed-contract evaluation enables more
controlled and interpretable comparison across wildfire tasks
and models.
We hope \ourfm\ and the fixed-contract framework provide a
foundation for future wildfire-specific Earth FMs, transfer
benchmarks, and decision-oriented evaluation protocols. 
More broadly, our research provides a reliable system to guide real-world intervention and resource allocation at the intersection of AI for environmental decision-making.

\paragraph{Limitations.} The conclusions apply to the task forms, scopes, evaluation rules, and comparator eligibility decisions used in this study.
The evaluation covers selected wildfire decision tasks and supporting retrieval and regression task forms.
They provide task-form evidence rather than a single score across all wildfire-related prediction tasks.

%% file: sections/appendix.tex
\appendix

\providecommand{\ms}[2]{\ensuremath{#1{\mkern1mu}_{\scriptscriptstyle \pm #2}}}

\section*{Appendix Contents}
\addcontentsline{toc}{section}{Appendix Contents}

\begin{center}
\begin{tabular}{@{}p{0.82\textwidth}r@{}}
\textbf{A\quad Evaluation Contract Specifications} & \pageref{sec:app_contract} \\
\quad A.1\enspace Matching Rule Definitions        & \pageref{sec:app_contract_matching} \\
\quad A.2\enspace Task-Form Contract Parameters    & \pageref{sec:app_contract_params} \\
\quad A.3\enspace Evaluation Scope Definitions     & \pageref{sec:app_contract_scope} \\[4pt]
\textbf{B\quad Controlled Check Details}           & \pageref{sec:app_checks} \\
\quad B.1\enspace Fixed-Output Check: Full Sweep   & \pageref{sec:app_checks_output} \\
\quad B.2\enspace Fixed-Feature Check: Selection Summary & \pageref{sec:app_checks_feature} \\
\quad B.3\enspace Selection Regret Under Matching Rules & \pageref{sec:app_checks_regret} \\
\quad B.4\enspace Additional Value Tables & \pageref{sec:app_checks_values} \\[4pt]
\textbf{C\quad Comparator Eligibility Notes}       & \pageref{sec:comparator_audit} \\[4pt]
\textbf{D\quad Seeded Audits}                      & \pageref{sec:app_seeded_audits} \\
\quad D.1\enspace Seed Robustness Summary          & \pageref{sec:app_seed_robustness} \\[4pt]
\textbf{E\quad Lightweight Head and Adaptation Details} & \pageref{sec:app_heads} \\[4pt]
\textbf{F\quad Limitations}                        & \pageref{sec:limitations} \\[4pt]
\textbf{G\quad Reproducibility and Evaluation Artifacts} & \pageref{sec:repro_compute_impact} \\
\end{tabular}
\end{center}

\noindent\textit{Retention rule.}
Appendix tables are retained when they add contract parameters, controlled-check arithmetic,
task-specific non-main metrics, seed summaries, eligibility checks, or protocol details.
Full task matrices and reference-summary tables that repeat the main result tables are not repeated here.

\clearpage

\section{Evaluation Contract Specifications}
\label{sec:app_contract}

\subsection{Matching Rule Definitions}
\label{sec:app_contract_matching}

The three matching rules used across occupancy task forms are defined as follows.

\noindent\textbf{Exact matching.}
A predicted unit-time pair $(i,t) \in \widehat{P}_\tau$ is counted as a true positive if and only if the same pair appears in the observed fire set $P = \{(i,t): y_{i,t}=1\}$.
This is the strictest rule and yields the lowest $F_1$ for any fixed output.

\noindent\textbf{Tolerated matching.}
A predicted pair $(i,t)$ is counted as correct if there exists an observed pair $(i',t') \in P$ such that $\|i - i'\|_\infty \le k$ and $|t - t'| \le \Delta t$, where $k$ is the spatial tolerance in grid cells and $\Delta t$ is the temporal tolerance in forecast steps.
Both parameters are fixed as part of the evaluation contract $\mathcal{C}$ before scoring.

\noindent\textbf{Union matching.}
A predicted pair is counted as a true positive if it satisfies either exact or tolerated matching.
The resulting union-$F_1$ provides an upper bound on decision performance under the chosen tolerance.

\noindent\textbf{Fixed parameter values.}
For occupancy, the spatial tolerance is $k=8$ grid cells.
The temporal tolerance is $\Delta t=3$ forecast steps for union matching and $\Delta t=0$ for spatial-only tolerance.
The threshold $\tau$ is selected on validation strict-$F_1$ before test scoring.
For fire spread, the spatial tolerance is $k=4$ grid cells, $\Delta t=0$, and the threshold is selected on validation spatial $F_1$.

\noindent Table~\ref{tab:app_matching_rule_params} records the fixed matching-rule parameters.

\begin{table}[h]
\centering
\small
\setlength{\tabcolsep}{10pt}
\renewcommand{\arraystretch}{1.2}
\caption{Matching-rule values used in the evaluation contracts.}
\label{tab:app_matching_rule_params}
\begin{tabular}{lll}
\toprule
\textbf{Parameter} & \textbf{Occupancy} & \textbf{Fire spread} \\
\midrule
\(k\) & 8 cells & 4 cells \\
\(\Delta t\) & 3 for union; 0 spatial-only & 0 \\
\(\tau\) & val. strict \(F_1\) & val. spatial \(F_1\) \\
\bottomrule
\end{tabular}
\end{table}

\subsection{Task-Form Contract Parameters}
\label{sec:app_contract_params}

Table~\ref{tab:app_contract_params_full} lists fixed scoring values not shown in the main contract map.

\begin{table}[h]
\centering
\scriptsize
\setlength{\tabcolsep}{3.5pt}
\renewcommand{\arraystretch}{1.2}
\caption{Fixed scoring values used by each task-form contract.}
\label{tab:app_contract_params_full}
\begin{adjustbox}{max width=\textwidth}
\begin{tabular}{llll}
\toprule
\textbf{\(\mathcal{T}\)} & \textbf{Scoring} & \textbf{Validation} & \textbf{\(\Omega\)} \\
\midrule
Occupancy & \(k=8,\Delta t=3\); exact/tol./union \(F_1\) & val. strict \(F_1\) & global; top-5/10/20\% fire-prone \\
Fire spread & \(k=4,\Delta t=0\); exact/spatial \(F_1\), AP & val. spatial \(F_1\) & spread-region cells \\
Final burned area & log-RMSE, log-MAE, Spearman \(\rho\) & val. log-RMSE & test events \\
Analog retrieval & nDCG@10; retrieved-event log error & val. nDCG@10 & test events \\
Smoke PM\(_{2.5}\) & RMSE, MAE, Pearson \(r\); exceedance 35 & val. RMSE & test stations \\
Extreme heat & RMSE-C, MAE-C, exceedance \(F_1\) & val. threshold 27/30/33\(^{\circ}\)C & heat-region stations \\
\bottomrule
\end{tabular}
\end{adjustbox}
\end{table}

\subsection{Evaluation Scope Definitions}
\label{sec:app_contract_scope}

\noindent\textbf{Global scope.}
Evaluation covers all spatial units in the domain, including fire-inactive regions.
This scope can mask model differences on fire-relevant locations because inactive cells inflate true-negative counts.

\noindent\textbf{Fire-prone scope.}
Evaluation is restricted to grid cells in the top-$k$\% of historical fire activity.
We report results for top-5\%, top-10\%, and top-20\% cutoffs.
The cutoff thresholds are derived from the training period and held fixed at test time.

\noindent\textbf{Spread region scope.}
For fire spread tasks, evaluation is restricted to the predicted and observed burned raster patches.
Only cells within the union of $\widehat{B}$ and $B$ contribute to metric computation.

\noindent\textbf{Fixed scope sizes.}
The global scope contains 8,085,000 test cells.
The fire-prone top-5\%, top-10\%, and top-20\% scopes contain 404,280, 808,560, and 1,617,000 test cells, respectively.
The spread-region scope is event-specific and uses the union of $\widehat{B}$ and $B$.

\begin{table}[h]
\centering
\small
\setlength{\tabcolsep}{8pt}
\renewcommand{\arraystretch}{1.2}
\caption{Scope values used in the evaluation contracts.}
\label{tab:app_scope_params}
\begin{tabular}{lcc}
\toprule
\textbf{\(\Omega\)} & \textbf{Definition} & \textbf{Units} \\
\midrule
Global & full domain & 8,085,000 test cells \\
Fire-prone top-5\% & top 5\% by training-period fire frequency & 404,280 test cells \\
Fire-prone top-10\% & top 10\% by training-period fire frequency & 808,560 test cells \\
Fire-prone top-20\% & top 20\% by training-period fire frequency & 1,617,000 test cells \\
Spread region & union of \(\widehat{B}\) and \(B\) & event-specific cells \\
\bottomrule
\end{tabular}
\end{table}

\clearpage

\section{Controlled Check Details}
\label{sec:app_checks}

\begin{figure}[t]
    \centering
    \includegraphics[width=\textwidth]{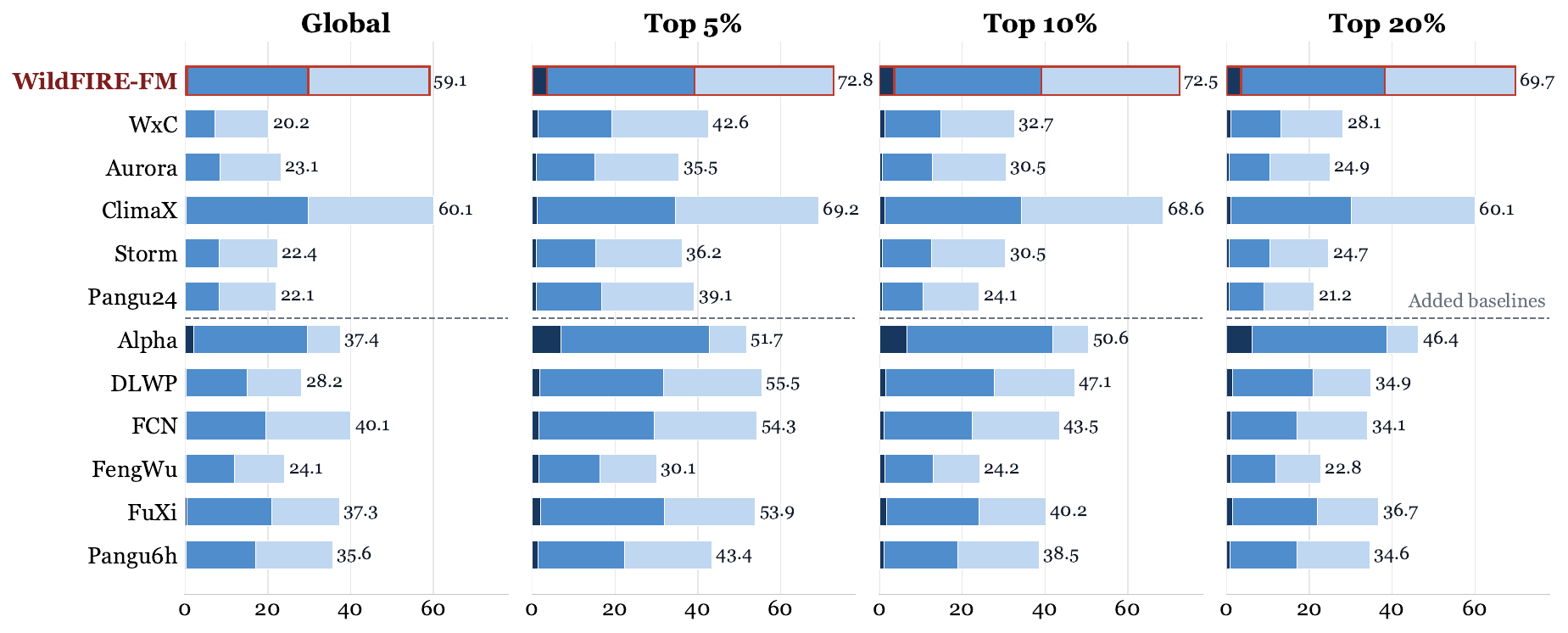}
    \caption{
\textbf{Matching-rule sensitivity in fire-prone occupancy (RQ1).}
Each row holds the score field \(S\), label field \(Y\), threshold, and \(\Omega\) fixed, and changes only \(\Lambda\).
Legend: \textcolor[HTML]{17375E}{$\blacksquare$} strict \(F_1\),
\textcolor[HTML]{4F8DCC}{$\blacksquare$} added \(F_1\) from spatial tolerance,
\textcolor[HTML]{BFD7F0}{$\blacksquare$} added \(F_1\) from union matching,
red outline \ourfm, and dashed line original weather FMs vs.\ added baselines.
The horizontal axis is \(F_1\) in percent.
}
    \label{fig:fireprone_contract_progression}
\end{figure}

\subsection{Fixed-Output Check: Full Sweep}
\label{sec:app_checks_output}

The fixed-output check holds the score field $S$ and label field $Y$ fixed and varies only $\Lambda$.
Table~\ref{tab:fireprone_contract_progression} reports the full global and fire-prone sweep for all retained backbones.
The same table is the numeric counterpart to Figure~\ref{fig:fireprone_contract_progression}.

\begin{table*}[t]
    \centering
    \scriptsize
    \setlength{\tabcolsep}{4pt}
    \caption{Occupancy \(F_1\) scores across global and fire-prone scopes. Global uses the full validation/test domain; top-\(k\) rows use train-defined fire-prone masks from historical fire frequency. Values are percentages from the same validation-selected strict threshold. Tolerance is spatial-only; union adds temporal and spatial matching. \(\Delta\) is union minus strict. Cells report five-seed mean with std in small type.}
    \label{tab:fireprone_contract_progression}
    \begin{tabular}{@{}llcccc@{}}
        \toprule
        Backbone & \(\Omega\) & Strict \(F_1\uparrow\) & Tol.\ \(F_1\uparrow\) & Union \(F_1\uparrow\) & \(\Delta\) \(\uparrow\) \\
        \midrule
        \ourfm & global & \ms{0.4546}{0.1412} & \ms{29.7484}{1.2868} & \ms{59.0656}{2.7372} & \ms{58.6109}{2.6945} \\
         & top 5\% & \ms{3.5604}{0.8809} & \ms{39.2617}{1.4011} & \ms{72.8280}{2.5784} & \ms{69.2676}{1.9960} \\
         & top 10\% & \ms{3.5575}{0.8799} & \ms{39.1665}{1.3906} & \ms{72.5204}{2.5670} & \ms{68.9629}{1.9888} \\
         & top 20\% & \ms{3.5300}{0.8700} & \ms{38.2849}{1.2952} & \ms{69.7228}{2.4664} & \ms{66.1928}{1.9273} \\
        \addlinespace[1pt]
        Prithvi-WxC & global & \ms{0.0552}{0.0039} & \ms{7.1649}{0.6557} & \ms{20.1853}{1.8299} & \ms{20.1301}{1.8297} \\
         & top 5\% & \ms{1.4119}{1.1635} & \ms{19.2636}{4.5019} & \ms{42.5793}{4.5495} & \ms{41.1674}{3.4846} \\
         & top 10\% & \ms{1.2376}{1.3201} & \ms{14.8780}{8.4429} & \ms{32.6913}{13.2085} & \ms{31.4536}{11.9053} \\
         & top 20\% & \ms{1.1520}{1.3770} & \ms{13.1512}{9.4556} & \ms{28.1319}{15.2866} & \ms{26.9800}{13.9224} \\
        \addlinespace[1pt]
        Aurora & global & \ms{0.0656}{0.0094} & \ms{8.5009}{1.9594} & \ms{23.1037}{4.9418} & \ms{23.0382}{4.9325} \\
         & top 5\% & \ms{0.9859}{0.9299} & \ms{15.1337}{6.0821} & \ms{35.4834}{11.0192} & \ms{34.4975}{10.3728} \\
         & top 10\% & \ms{0.7790}{1.0453} & \ms{12.7381}{6.5558} & \ms{30.5305}{10.8842} & \ms{29.7515}{9.8656} \\
         & top 20\% & \ms{0.6655}{1.1043} & \ms{10.5304}{7.4309} & \ms{24.9444}{12.5844} & \ms{24.2790}{11.4943} \\
        \addlinespace[1pt]
        ClimaX & global & \ms{0.3480}{0.0754} & \ms{29.7535}{3.6073} & \ms{60.1506}{7.5865} & \ms{59.8026}{7.5454} \\
         & top 5\% & \ms{1.2937}{0.1086} & \ms{34.5791}{2.3772} & \ms{69.2186}{5.7215} & \ms{67.9249}{5.7263} \\
         & top 10\% & \ms{1.2522}{0.1602} & \ms{34.3341}{2.2852} & \ms{68.5713}{5.5377} & \ms{67.3191}{5.5538} \\
         & top 20\% & \ms{1.0287}{0.2686} & \ms{30.2140}{4.2857} & \ms{60.0650}{7.5674} & \ms{59.0363}{7.5891} \\
        \addlinespace[1pt]
        StormCast & global & \ms{0.0626}{0.0119} & \ms{8.1951}{2.1895} & \ms{22.3817}{5.4294} & \ms{22.3191}{5.4178} \\
         & top 5\% & \ms{0.9573}{0.8011} & \ms{15.3219}{5.5337} & \ms{36.1857}{9.7331} & \ms{35.2284}{9.1816} \\
         & top 10\% & \ms{0.7284}{0.9280} & \ms{12.6669}{6.3290} & \ms{30.4748}{10.6527} & \ms{29.7464}{9.7494} \\
         & top 20\% & \ms{0.5795}{0.9104} & \ms{10.4157}{7.3437} & \ms{24.6598}{12.3973} & \ms{24.0803}{11.4988} \\
        \addlinespace[1pt]
        DLWP & global & \ms{0.1693}{0.0419} & \ms{14.9148}{3.2446} & \ms{28.1901}{6.9658} & \ms{28.0208}{6.9257} \\
         & top 5\% & \ms{1.8054}{0.4835} & \ms{31.7231}{3.2923} & \ms{55.4596}{5.2920} & \ms{53.6542}{5.4752} \\
         & top 10\% & \ms{1.6110}{0.5999} & \ms{27.6581}{5.9216} & \ms{47.1269}{8.0111} & \ms{45.5158}{7.7927} \\
         & top 20\% & \ms{1.5248}{0.8987} & \ms{20.9403}{4.7971} & \ms{34.9301}{7.8471} & \ms{33.4054}{7.8760} \\
        \addlinespace[1pt]
        FCN & global & \ms{0.2829}{0.0839} & \ms{19.5061}{3.3412} & \ms{40.0604}{9.3701} & \ms{39.7775}{9.3423} \\
         & top 5\% & \ms{1.6231}{0.5064} & \ms{29.3769}{2.7626} & \ms{54.3033}{7.4089} & \ms{52.6801}{7.4389} \\
         & top 10\% & \ms{1.1777}{0.5118} & \ms{22.4217}{3.9803} & \ms{43.4510}{9.2513} & \ms{42.2734}{9.0251} \\
         & top 20\% & \ms{0.9962}{0.4315} & \ms{16.9792}{3.9371} & \ms{34.0859}{8.2616} & \ms{33.0897}{7.9275} \\
        \addlinespace[1pt]
        FengWu & global & \ms{0.2613}{0.0757} & \ms{12.0050}{6.0239} & \ms{24.1022}{13.6293} & \ms{23.8410}{13.5736} \\
         & top 5\% & \ms{1.5695}{0.3592} & \ms{16.2763}{3.7024} & \ms{30.1055}{5.0103} & \ms{28.5360}{4.7696} \\
         & top 10\% & \ms{1.2427}{0.5333} & \ms{12.9503}{5.6052} & \ms{24.1854}{8.6854} & \ms{22.9427}{8.1863} \\
         & top 20\% & \ms{1.1192}{0.5023} & \ms{11.9508}{5.0745} & \ms{22.7860}{7.9115} & \ms{21.6668}{7.4438} \\
        \addlinespace[1pt]
        FuXi & global & \ms{0.3774}{0.1212} & \ms{21.0323}{4.8211} & \ms{37.2888}{9.4470} & \ms{36.9114}{9.4327} \\
         & top 5\% & \ms{2.0307}{0.6800} & \ms{31.8944}{4.7331} & \ms{53.9308}{8.3822} & \ms{51.9001}{8.6878} \\
         & top 10\% & \ms{1.6542}{0.7316} & \ms{24.0128}{5.7784} & \ms{40.2140}{9.9307} & \ms{38.5597}{9.7744} \\
         & top 20\% & \ms{1.3646}{0.6773} & \ms{21.9548}{5.8601} & \ms{36.7314}{10.0289} & \ms{35.3668}{9.9223} \\
        \addlinespace[1pt]
        Pangu-Weather & global & \ms{0.2755}{0.1089} & \ms{17.0909}{4.0477} & \ms{35.6386}{9.0327} & \ms{35.3630}{9.0774} \\
         & top 5\% & \ms{1.3656}{0.3064} & \ms{22.2222}{6.8613} & \ms{43.4234}{13.2383} & \ms{42.0578}{13.0599} \\
         & top 10\% & \ms{1.0931}{0.3535} & \ms{18.9337}{5.9329} & \ms{38.5325}{11.7221} & \ms{37.4394}{11.5261} \\
         & top 20\% & \ms{0.8844}{0.3601} & \ms{17.0172}{5.4859} & \ms{34.5688}{10.2932} & \ms{33.6844}{10.1334} \\
        \addlinespace[1pt]
        AlphaEarth & global & \ms{2.0606}{0.4404} & \ms{29.4476}{6.0064} & \ms{37.4286}{9.9458} & \ms{35.3679}{10.0271} \\
         & top 5\% & \ms{6.9133}{0.8450} & \ms{42.8790}{4.6087} & \ms{51.7449}{8.7321} & \ms{44.8315}{9.0763} \\
         & top 10\% & \ms{6.6366}{0.9901} & \ms{41.8981}{5.9454} & \ms{50.5712}{10.0057} & \ms{43.9346}{9.9156} \\
         & top 20\% & \ms{6.1908}{1.1330} & \ms{38.8325}{7.4966} & \ms{46.3833}{12.1697} & \ms{40.1925}{11.6788} \\
        \bottomrule
    \end{tabular}
\end{table*}

\subsection{Fixed-Feature Check: Selection Summary}
\label{sec:app_checks_feature}

The paper appendix keeps the fixed-feature result at the selection-summary level.
The full per-head rows are retained in the supplementary CSV files and are not repeated as a manuscript table because many degenerate heads produce identical zero decision scores.
The supplementary selection rows report the decision-score loss after changing only the head-selection metric.

\subsection{Selection Regret Under Matching Rules}
\label{sec:app_checks_regret}

The fixed-feature check trains the same head family $\mathcal{H}$ on a fixed feature source and changes only the selection metric.
Table~\ref{tab:appendix_selection_regret_tolerance} reports the same selection comparison under exact, tolerated, and union matching.
Here, \(h_R\) is selected by PR-AUC and \(h_D\) is selected by the decision metric.
The reported regret is \(D(h_D)-D(h_R)\).
Exact zero entries mean the two selectors give the same decision score for all five seeds.

\begin{table*}[!t]
    \centering
    \scriptsize
    \setlength{\tabcolsep}{4pt}
    \caption{Selection-regret values under exact, tolerated, and union matching. Values are percentage-point regret from selecting \(h_R\) by PR-AUC instead of \(h_D\) by the decision metric. Rows report mean with small std over five seeds; \(0.0000\) denotes exact zero regret.}
    \label{tab:appendix_selection_regret_tolerance}
    \begin{adjustbox}{max width=\textwidth}
    \begin{tabular}{llccc}
        \toprule
        \textbf{Feature} & \textbf{\(\Omega\)} & \textbf{Exact regret} & \textbf{Tolerated regret} & \textbf{Union regret} \\
        \midrule
        \ourfm & global & 0.0000 & \ms{8.7830}{9.6705} & \ms{8.7830}{9.6705} \\
        \ourfm & fire-prone & 0.0000 & \ms{3.4027}{3.2045} & \ms{3.4027}{3.2045} \\
        Prithvi-WxC & global & 0.0000 & 0.0000 & 0.0000 \\
        Prithvi-WxC & fire-prone & 0.0000 & 0.0000 & 0.0000 \\
        Aurora & global & \ms{0.0200}{0.0267} & \ms{9.8520}{12.9878} & \ms{9.8520}{12.9878} \\
        Aurora & fire-prone & \ms{0.8203}{1.8341} & \ms{14.3919}{32.1219} & \ms{14.3919}{32.1219} \\
        ClimaX & global & \ms{0.0003}{0.0004} & \ms{0.1296}{0.1775} & \ms{0.1296}{0.1775} \\
        ClimaX & fire-prone & 0.0000 & 0.0000 & 0.0000 \\
        StormCast & global & 0.0000 & 0.0000 & 0.0000 \\
        StormCast & fire-prone & 0.0000 & 0.0000 & 0.0000 \\
        DLWP & global & 0.0000 & 0.0000 & 0.0000 \\
        DLWP & fire-prone & \ms{0.0770}{0.1100} & \ms{4.3266}{4.3323} & \ms{4.3266}{4.3323} \\
        FCN & global & 0.0000 & 0.0000 & 0.0000 \\
        FCN & fire-prone & \ms{0.0006}{0.0013} & \ms{1.1680}{1.9872} & \ms{1.1680}{1.9872} \\
        FengWu & global & 0.0000 & 0.0000 & 0.0000 \\
        FengWu & fire-prone & \ms{0.0691}{0.1191} & \ms{0.5222}{0.6239} & \ms{0.5222}{0.6239} \\
        FuXi & global & 0.0000 & 0.0000 & 0.0000 \\
        FuXi & fire-prone & 0.0000 & \ms{0.1084}{0.1729} & \ms{0.1084}{0.1729} \\
        Pangu-Weather & global & 0.0000 & 0.0000 & 0.0000 \\
        Pangu-Weather & fire-prone & \ms{0.0728}{0.1179} & \ms{0.1849}{0.3263} & \ms{0.1849}{0.3263} \\
        AlphaEarth & global & 0.0000 & \ms{17.2217}{8.8492} & \ms{17.2217}{8.8492} \\
        AlphaEarth & fire-prone & 0.0000 & \ms{3.8804}{5.9483} & \ms{3.8804}{5.9483} \\
        \bottomrule
    \end{tabular}
    \end{adjustbox}
\end{table*}

\subsection{Additional Value Tables}
\label{sec:app_checks_values}

Table~\ref{tab:app_occupancy_ppr_scope}
reports the predicted-positive rate behind the occupancy \(F_1\) sweep.

\begin{table*}[t]
\centering
\small
\setlength{\tabcolsep}{4pt}
\renewcommand{\arraystretch}{1.18}
\caption{For fixed occupancy \(\mathcal{T}\), this table reports predicted-positive rate.
Values are percentages under the same validation-selected strict threshold.
Scopes \(\Omega\) are fixed before test scoring; cells report five-seed mean with std in small type.}
\label{tab:app_occupancy_ppr_scope}
\begin{tabular}{lcccc}
\toprule
\textbf{Backbone} & \textbf{\(\Omega=\)global} & \textbf{\(\Omega=\)top 5\%} & \textbf{\(\Omega=\)top 10\%} & \textbf{\(\Omega=\)top 20\%} \\
\midrule
\ourfm & \ms{1.6808}{0.3684} & \ms{3.0619}{1.0925} & \ms{1.5310}{0.5463} & \ms{0.7655}{0.2732} \\
Prithvi-WxC & \ms{61.9711}{30.9101} & \ms{57.4117}{47.8987} & \ms{58.4565}{51.0897} & \ms{58.9788}{52.6991} \\
Aurora & \ms{55.5849}{19.7524} & \ms{57.2238}{35.3400} & \ms{68.7942}{37.6958} & \ms{67.2891}{38.3991} \\
ClimaX & \ms{5.6763}{3.9261} & \ms{24.0091}{9.2816} & \ms{11.8450}{4.5067} & \ms{5.7442}{4.1341} \\
StormCast & \ms{60.6507}{17.4895} & \ms{57.6017}{35.2921} & \ms{68.0766}{37.3899} & \ms{67.8397}{39.2410} \\
DLWP & \ms{4.3221}{1.5619} & \ms{9.4001}{5.0807} & \ms{4.9700}{3.6849} & \ms{1.9198}{1.4678} \\
FCN & \ms{1.5202}{1.3446} & \ms{4.7856}{2.9409} & \ms{2.7257}{1.6353} & \ms{0.8368}{0.2358} \\
FengWu & \ms{0.4277}{0.4830} & \ms{0.6004}{0.3041} & \ms{0.2609}{0.1935} & \ms{0.1501}{0.1206} \\
FuXi & \ms{0.4505}{0.2773} & \ms{2.9315}{2.6392} & \ms{0.5197}{0.6074} & \ms{0.3621}{0.4346} \\
Pangu-Weather & \ms{1.0801}{1.1308} & \ms{2.0549}{2.1893} & \ms{1.4029}{1.4739} & \ms{1.0103}{1.1084} \\
AlphaEarth & \ms{0.0691}{0.0499} & \ms{0.2826}{0.1497} & \ms{0.1524}{0.0770} & \ms{0.0656}{0.0414} \\
\bottomrule
\end{tabular}
\end{table*}

Tables~\ref{tab:app_spread_ap_by_scope}--\ref{tab:app_heat_event_pr}
report additional values that are not repeated in the main tables.
Each table fixes the task \(\mathcal{T}\) and reports either a different \(\Omega\), metric, or event subset.

\begin{table*}[t]
\centering
\scriptsize
\setlength{\tabcolsep}{3pt}
\caption{For fixed spread \(\mathcal{T}\) and strict \(\Lambda\), this table reports AP under three \(\Omega\) scopes: full test, top-5\% train-fire area, and top-10\% train-fire area. Values are percentages; cells report mean with small std.}
\label{tab:app_spread_ap_by_scope}
\begin{tabular}{lccc}
\toprule
Backbone & full \(\Omega\) AP & top-5\% \(\Omega\) AP & top-10\% \(\Omega\) AP \\
\midrule
\ourfm & \ms{30.0197}{1.5651} & \ms{40.7452}{2.0542} & \ms{37.4096}{1.8731} \\
Prithvi-WxC & \ms{4.8319}{0.1731} & \ms{12.6086}{0.4468} & \ms{8.7051}{0.1889} \\
Aurora & \ms{17.7723}{0.4293} & \ms{30.3106}{0.9404} & \ms{26.4732}{0.6932} \\
ClimaX & \ms{11.1726}{0.2337} & \ms{25.7871}{1.2896} & \ms{19.9977}{1.2217} \\
StormCast & \ms{8.1147}{1.1569} & \ms{18.5461}{1.1727} & \ms{14.1286}{1.2956} \\
DLWP & \ms{9.2142}{2.6587} & \ms{19.3346}{2.3922} & \ms{14.9788}{2.6696} \\
FCN & \ms{6.6774}{1.3001} & \ms{16.7396}{3.2955} & \ms{11.9308}{2.3881} \\
FengWu & \ms{11.0046}{2.7092} & \ms{21.1506}{1.2163} & \ms{17.0113}{1.5778} \\
FuXi & \ms{13.5507}{0.3840} & \ms{22.5434}{0.4100} & \ms{19.1964}{0.3943} \\
Pangu-Weather & \ms{10.6250}{1.4643} & \ms{19.8294}{1.3044} & \ms{15.8013}{1.1602} \\
AlphaEarth & \ms{12.2847}{1.3562} & \ms{22.8692}{0.4915} & \ms{18.2992}{1.2110} \\
\bottomrule
\end{tabular}
\end{table*}

\begin{table*}[t]
\centering
\scriptsize
\setlength{\tabcolsep}{3pt}
\caption{For fixed final-area \(\mathcal{T}\) and \(\Omega\), this table reports median log error and acre-scale errors in addition to the main log-RMSE/log-MAE/Spearman metrics. Cells report mean with small std.}
\label{tab:app_burned_area_median_acre}
\begin{tabular}{lccc}
\toprule
Backbone & log median AE & acre median AE & acre MAPE \\
\midrule
\ourfm & \ms{1.0235}{0.0982} & \ms{4504.0692}{459.0483} & \ms{1.4525}{0.0254} \\
Prithvi-WxC & \ms{1.2184}{0.2107} & \ms{5375.8770}{788.7906} & \ms{1.9517}{0.2875} \\
Aurora & \ms{1.4547}{0.0301} & \ms{9904.9483}{457.4260} & \ms{6.8728}{3.0026} \\
ClimaX & \ms{1.6841}{0.1818} & \ms{18130.4820}{3248.3873} & \ms{8.2373}{2.8540} \\
StormCast & \ms{1.4522}{0.1519} & \ms{11155.7881}{2020.8656} & \ms{4.6142}{1.1500} \\
DLWP & \ms{1.0952}{0.1306} & \ms{4406.9315}{303.0944} & \ms{1.7357}{0.3625} \\
FCN & \ms{1.1688}{0.1139} & \ms{5166.9993}{213.0333} & \ms{2.0800}{0.4004} \\
FengWu & \ms{1.1589}{0.1772} & \ms{5137.2822}{628.7543} & \ms{2.0944}{0.4545} \\
FuXi & \ms{1.1855}{0.0612} & \ms{5697.7117}{796.8785} & \ms{2.4411}{0.5567} \\
Pangu-Weather & \ms{1.1221}{0.1470} & \ms{5092.3621}{483.8243} & \ms{1.9571}{0.3113} \\
AlphaEarth & \ms{1.7459}{0.6057} & \ms{15110.7573}{7106.3417} & \ms{9.7398}{2.7425} \\
\bottomrule
\end{tabular}
\end{table*}

\begin{table*}[t]
\centering
\scriptsize
\setlength{\tabcolsep}{3pt}
\caption{For fixed retrieval \(\mathcal{T}\) and \(\Omega\), this table reports nDCG@5, best log gap, and rank \(\rho\) in addition to the main nDCG@10/log-error metrics. Cells report mean with small std.}
\label{tab:app_analog_rank_depth}
\begin{tabular}{lccc}
\toprule
Backbone & nDCG@5 & best log gap & rank $\rho$ \\
\midrule
\ourfm & \ms{0.5175}{0.0445} & \ms{0.1868}{0.0285} & \ms{0.6019}{0.1460} \\
Prithvi-WxC & \ms{0.3591}{0.0107} & \ms{0.2151}{0.0594} & \ms{0.1514}{0.1489} \\
Aurora & \ms{0.4423}{0.0210} & \ms{0.1551}{0.0437} & \ms{0.2162}{0.1856} \\
ClimaX & \ms{0.4151}{0.0293} & \ms{0.2129}{0.0653} & \ms{0.1587}{0.2831} \\
StormCast & \ms{0.3960}{0.0240} & \ms{0.1714}{0.0310} & \ms{0.1258}{0.1625} \\
DLWP & \ms{0.3795}{0.0274} & \ms{0.1944}{0.0807} & \ms{-0.3865}{0.2802} \\
FCN & \ms{0.4250}{0.0112} & \ms{0.1856}{0.0846} & \ms{-0.1357}{0.2571} \\
FengWu & \ms{0.4228}{0.0310} & \ms{0.1870}{0.0858} & \ms{-0.1926}{0.2194} \\
FuXi & \ms{0.4544}{0.0356} & \ms{0.2171}{0.0806} & \ms{-0.1367}{0.2885} \\
Pangu-Weather & \ms{0.3988}{0.0506} & \ms{0.1901}{0.0838} & \ms{-0.1970}{0.2216} \\
AlphaEarth & \ms{0.5276}{0.0531} & \ms{0.1782}{0.0454} & \ms{0.4639}{0.2802} \\
\bottomrule
\end{tabular}
\end{table*}

\begin{table*}[t]
\centering
\scriptsize
\setlength{\tabcolsep}{3pt}
\caption{For fixed smoke \(\mathcal{T}\) and station \(\Omega\), this table reports RMSE, MAE, and 90th-percentile absolute error on test rows with observed PM$_{2.5}\ge35$; std uses a row bootstrap over those rows. Cells report mean with small std.}
\label{tab:app_smoke_high_event}
\begin{tabular}{lccc}
\toprule
Backbone & high-smoke RMSE & high-smoke MAE & high-smoke 90th AE \\
\midrule
\ourfm & \ms{47.4870}{0.6346} & \ms{34.3954}{0.7654} & \ms{65.6213}{3.8778} \\
Prithvi-WxC & \ms{57.2224}{1.7268} & \ms{47.3871}{0.3153} & \ms{74.9666}{3.2381} \\
Aurora & \ms{57.2752}{1.7248} & \ms{47.4368}{0.3149} & \ms{75.0755}{3.1074} \\
ClimaX & \ms{57.2828}{1.7239} & \ms{47.4407}{0.3140} & \ms{75.1012}{3.0777} \\
StormCast & \ms{56.6512}{1.7517} & \ms{46.7914}{0.3281} & \ms{74.0794}{3.4707} \\
DLWP & \ms{57.0075}{1.7359} & \ms{47.1971}{0.3198} & \ms{74.4936}{3.3826} \\
FCN & \ms{57.0582}{1.7339} & \ms{47.2401}{0.3187} & \ms{74.6431}{3.1982} \\
FengWu & \ms{57.0158}{1.7357} & \ms{47.1957}{0.3194} & \ms{74.5652}{3.2871} \\
FuXi & \ms{56.9622}{1.7371} & \ms{47.1508}{0.3201} & \ms{74.3278}{3.4435} \\
Pangu-Weather & \ms{57.1282}{1.7307} & \ms{47.3050}{0.3170} & \ms{74.6830}{3.2375} \\
AlphaEarth & \ms{48.0665}{0.7904} & \ms{35.6088}{0.7341} & \ms{66.7613}{3.9235} \\
\bottomrule
\end{tabular}
\end{table*}

\begin{table*}[t]
\centering
\scriptsize
\setlength{\tabcolsep}{3pt}
\caption{For fixed heat \(\mathcal{T}\) and heat-region \(\Omega\), this table reports precision and recall for the exceedance label used by the main \(F_1\). Cells report mean with small std.}
\label{tab:app_heat_event_pr}
\begin{tabular}{lcc}
\toprule
Backbone & precision & recall \\
\midrule
\ourfm & \ms{0.9767}{0.0117} & \ms{0.9330}{0.0299} \\
Prithvi-WxC & \ms{0.8260}{0.0030} & \ms{0.9173}{0.0033} \\
Aurora & \ms{0.5920}{0.0347} & \ms{0.0517}{0.0020} \\
ClimaX & \ms{0.7397}{0.0099} & \ms{0.7994}{0.0051} \\
StormCast & \ms{0.8840}{0.0237} & \ms{0.9320}{0.0165} \\
DLWP & \ms{0.9429}{0.0085} & \ms{0.8899}{0.0167} \\
FCN & \ms{0.9408}{0.0097} & \ms{0.9111}{0.0127} \\
FengWu & \ms{0.3808}{0.2719} & \ms{0.0266}{0.0267} \\
FuXi & \ms{0.3262}{0.1262} & \ms{0.1810}{0.0481} \\
Pangu-Weather & \ms{0.1159}{0.0743} & \ms{0.0112}{0.0032} \\
AlphaEarth & \ms{0.9824}{0.0040} & \ms{0.9278}{0.0178} \\
\bottomrule
\end{tabular}
\end{table*}

\clearpage

\section{Comparator Eligibility Notes}
\label{sec:comparator_audit}

All numeric comparator rows in Tables~\ref{tab:primary_results} and~\ref{tab:supporting_results}
are included only after the task form, metric, matching rule, scope, and head family are fixed.
The appendix does not repeat those full matrices.
The key eligibility rule is simple: reported rows satisfy the same contract as the row block in which they appear, while excluded rows are excluded because their representation or output form does not satisfy that contract.

\noindent\textbf{Reading rule.}
Exact-only, tolerated, union, ranking, retrieval, and regression scores answer different questions.
The fixed-contract reading is therefore to compare entries only within one row block and not to average across task forms.

\clearpage

\section{Seeded Audits}
\label{sec:app_seeded_audits}

\subsection{Seed Robustness Summary}
\label{sec:app_seed_robustness}

Table~\ref{tab:app_seed_robustness} summarizes stochastic checks used to support the reported mean-with-std convention.
It is not a replacement for the main fixed-contract result tables.

\begin{table}[h]
\centering
\small
\setlength{\tabcolsep}{5pt}
\renewcommand{\arraystretch}{1.2}
\caption{Seed summaries for stochastic checks. Values report mean with small std over completed seeds.}
\label{tab:app_seed_robustness}
\begin{adjustbox}{max width=\textwidth}
\begin{tabular}{p{0.28\textwidth}cllp{0.18\textwidth}}
\toprule
\textbf{\(\mathcal{T}\) check} & \textbf{Seeds} & \textbf{Primary value} & \textbf{Other value(s)} & \textbf{Reading} \\
\midrule
Final burned area &
5 & log-RMSE \ms{1.1657}{0.0126} &
log-MAE \ms{1.0423}{0.0081}; Spear.\ \ms{0.6298}{0.0338} &
stable across seeds \\
Smoke PM\(_{2.5}\) &
5 & RMSE \ms{4.4646}{0.0060} &
MAE \ms{2.4108}{0.0016}; \(r\) \ms{0.6368}{0.0013} &
stable at table precision \\
Extreme heat &
5 & RMSE-C \ms{0.2179}{0.0043} &
MAE-C \ms{0.1787}{0.0018}; exceed.\ \(F_1\) \ms{0.9541}{0.0164} &
stable across seeds \\
Fire spread &
5 & exact \(F_1\) \ms{37.6700}{0.9800} &
spatial \(F_1\) \ms{80.9700}{2.0200}; AP \ms{30.0900}{1.2500} &
stable across seeds \\
Aurora paired-head check &
5 & fire-prone score diff.\ \ms{6.3500}{13.2800} &
PR-AUC and union choices differ in 2/5 seeds &
variable across seeds \\
\bottomrule
\end{tabular}
\end{adjustbox}
\end{table}

\clearpage

\section{Lightweight Head and Adaptation Details}
\label{sec:app_heads}

All frozen-transfer comparisons use the same five lightweight head architectures applied
on top of the frozen backbone representations.
Table~\ref{tab:app_head_architectures} summarises each head family, its architecture,
approximate parameter count, and the adaptation procedure used.

\begin{table}[h]
\centering
\small
\setlength{\tabcolsep}{5pt}
\renewcommand{\arraystretch}{1.3}
\caption{Lightweight head architectures used in the fixed-contract transfer comparisons.
All heads are trained from random initialisation on the frozen backbone features.
Parameter counts are approximate and depend on the feature dimensionality of each backbone.}
\label{tab:app_head_architectures}
\begin{tabular}{p{0.15\textwidth}p{0.30\textwidth}p{0.12\textwidth}p{0.33\textwidth}}
\toprule
\textbf{$\mathcal{A}$ head} & \textbf{Architecture} & \textbf{Approx.\ params} & \textbf{Notes} \\
\midrule
Constant prior &
  Outputs a fixed bias vector, ignoring input features. &
  Output dimension only &
  Provides a degenerate baseline; selected when backbone features carry no useful signal. \\
Linear probe &
  Single linear layer mapping backbone features to output. No nonlinearity. &
  $d\times c + c$ &
  Standard frozen-representation baseline. \\
Pixel MLP &
  Two-layer MLP applied independently per spatial unit. &
  $d\times h + h\times c$ &
  Captures per-pixel nonlinearity; ignores spatial context. \\
Shallow adapter &
  Two-layer MLP with a spatial context window; uses $3\times3$ convolution before the linear output. &
  $9dh + hc$ &
  Balances local spatial context with parameter efficiency. \\
Wide adapter &
  Shallow adapter with wider hidden dimension. &
  $9dH + Hc$ &
  Higher capacity variant; can overfit on small fire-event sets. \\
\bottomrule
\end{tabular}
\end{table}

\noindent\textbf{Training protocol.}
Each occupancy head-control run uses seeds $\{1,7,42,99,123\}$, the five heads listed above, and the fixed variants identity, erode-r1, and close-r1.
The spread U-Net reference is trained for 4 epochs.
The threshold $\tau$ is selected on the validation split by maximising union-$F_1$ (for occupancy) or spatial $F_1$ (for spread) and held fixed at test time.
Morphology parameters (spatial tolerance $k$, temporal tolerance $\Delta t$) are fixed as part of the evaluation contract and are not tuned after validation.

\noindent\textbf{Head selection procedure.}
For each (feature source, scope, seed) tuple, all five heads are trained independently.
The PR-AUC-based selector picks $h_R = \arg\max_{h \in \mathcal{H}} R(h)$ on the validation set;
the decision-based selector picks $h_D = \arg\max_{h \in \mathcal{H}} D(h)$ on the same set.
The selection regret $\delta = D(h_D) - D(h_R) \ge 0$ is computed on the held-out test set.

\clearpage

\section{Limitations}
\label{sec:limitations}

The conclusions apply to the task forms, scopes, evaluation rules, and comparator eligibility decisions used in this study.
The evaluation covers selected wildfire decision tasks and supporting retrieval and regression task forms.
Comparator eligibility is fixed before metric values are interpreted.
This eligibility rule keeps each comparison within one task-form contract.
It also leaves some model and task pairs outside the evaluated comparison set by design.

The transfer comparison uses frozen backbones with lightweight heads.
The results therefore describe frozen-backbone transfer under the allowed head families in each contract.
Full fine-tuning, alternative adaptation procedures, and broader head families are outside the evaluated scope.
The task-specific reference baselines serve as empirical anchors for same-contract comparison.
\ourfm is a regional wildfire reference for the reported California fixed-contract experiments.

The supporting retrieval and regression checks bound the primary spatial decision claim.
They provide task-form evidence rather than a single score across all wildfire-related prediction tasks.
The analysis focuses on the reported metric families, matching rules, and fixed comparison choices.
Operational response rules, intervention costs, and deployment policies are part of wildfire early-warning use contexts~\cite{goldammer1999early,pickell2017early,farahmand2020fdeo}.
They are outside the scope of this evaluation study and are not inferred from the reported scores.

\clearpage

\section{Reproducibility and Evaluation Artifacts}
\label{sec:repro_compute_impact}
\subsection{External Assets and Terms of Use}
\label{sec:external_assets_terms}

We use external datasets and model assets only for research evaluation.
Access to each asset follows the original provider's portal, license, or terms of use; this submission does not imply that all assets are openly redistributable.
We do not redistribute raw external datasets, provider-hosted embeddings, or third-party model weights.
Table~\ref{tab:external_assets_licenses} records the source and terms-of-use status used to interpret reproducibility.

\begin{table}[h]
\centering
\small
\setlength{\tabcolsep}{4pt}
\renewcommand{\arraystretch}{1.18}
\caption{External assets used by the study and their source or terms-of-use status.}
\label{tab:external_assets_licenses}
\begin{tabular}{p{0.25\textwidth}p{0.34\textwidth}p{0.34\textwidth}}
\toprule
\textbf{Asset family} & \textbf{Use in this study} & \textbf{Source and terms-of-use note} \\
\midrule
NOAA HRRR fields~\cite{noaa_hrrr_ncei,noaa_hrrr_emc}
& Dynamic weather inputs for \ourfm and transfer tasks.
& NOAA provider terms and citation requirements apply. \\
NASA FIRMS~\cite{nasa_firms}
& Active-fire occupancy supervision.
& NASA Earthdata/FIRMS access terms and citation requirements apply. \\
LANDFIRE and WRC layers~\cite{landfire_fbfm40,landfire_canopy_cover,usfs_wrc_housing_density}
& Static fuel, canopy, and exposure context.
& Original geospatial-product provider terms and citations apply. \\
LandScan~\cite{ornl_landscan_2024}
& Static population context.
& ORNL/LandScan source-specific access terms apply; raw data are not redistributed. \\
WFIGS and MTBS~\cite{nifc_wfigs_perimeters,mtbs_usgs_2025}
& Event-level resources for burned-area and analog tasks.
& Original incident/perimeter-product provider terms and citations apply. \\
External Earth-FM baselines~\cite{schmude2024prithviwxc,bodnar2025aurora,nguyen2023climax,pathak2024stormcast,weyn2020dlwp,pathak2022fourcastnet,chen2023fengwu,chen2023fuxi,bi2023panguweather,brown2025alphaearth}
& Frozen comparator representations or task-model baselines.
& Original model-provider licenses and access terms apply; third-party weights are not redistributed. \\
\bottomrule
\end{tabular}
\end{table}

This note supports the NeurIPS checklist and identifies the files that support the reported claims.
This file statement does not imply full raw-data release.
The main claims can be checked from the manuscript contracts, metric
definitions, and per-head result files, even if full raw-data release is
delayed or limited. Sections~3 and~4 specify the contract components used by
the main claims: task definition, split logic, label space, tolerance
parameters, scope definitions, threshold or operating-point rules, and
lightweight-head set.

The supplementary source includes the check scripts, per-head and per-seed
CSV result files, and \LaTeX{} result tables for the expanded check and matching-rule support.
These files expose exact \(F_1\),
tolerated \(F_1\), union-\(F_1\), PR-AUC, per-head selection,
top-1 agreement, and selection-regret arithmetic. The manuscript also includes
full figure and table reproduction values in result tables and appendix tables.
These files provide a runnable check of the
selection-regret arithmetic and the table-construction logic from fixed
per-head rows. The seeded occupancy check uses seeds
$\{1,7,42,99,123\}$, and the spread task-specific U-Net check uses repeated seeds; reported error bars are standard deviations over the completed
seeded runs. Full raw wildfire inputs and large feature arrays are not
released at submission because redistribution and storage constraints require a
separate review.

For stochastic results, the paper reports mean with standard deviation over repeated seeds.
For fixed-output or fixed-feature controls, the table uses one fixed output or feature set; the changed item is the matching rule or selection metric.

The reported experiments use two resource classes on a shared Slurm-managed
cluster. Tabular retrieval/regression checks and same-feature head controls run
on CPU workers with 4 to 8 cores, 24 to 64~GB host memory, and 2 to 4~hour wall-clock
limits. Spread U-Net training and threshold calibration run on single-GPU jobs
with one B200 GPU, 8 CPU cores, 96~GB host memory, and a 4~hour wall-clock
limit. The seed/check waves reported in the appendix correspond to roughly
78 CPU job-hours and 12 GPU job-hours of scheduled wall-clock budget;
exploratory runs are not included in the reported compute accounting.

The raw-data limitation is separate from the selection-regret files.
The supplementary source is sufficient to inspect the selection-regret arithmetic and reproduce the reported tables.
Full end-to-end recomputation from raw wildfire inputs is not included at submission because redistribution review is still required.
The broader impact is evaluation-facing rather than operational.
Better reading of wildfire transfer evidence can reduce overconfident benchmark claims, while misread transfer results could still encourage inappropriate reliance on models with low decision scores.
For that reason, the paper keeps its claims wildfire-centered, decision-task
specific, and explicitly separate from any predictive deployment
recommendation.